\documentclass{article} % For LaTeX2e
\usepackage{iclr2025_conference,times}

\usepackage{amsmath,amsfonts,bm}

\def\eqref#1{equation~\ref{#1}}
\def\1{\bm{1}}

\def\vt{{\bm{t}}}

\DeclareMathAlphabet{\mathsfit}{\encodingdefault}{\sfdefault}{m}{sl}
\SetMathAlphabet{\mathsfit}{bold}{\encodingdefault}{\sfdefault}{bx}{n}

\newcommand{\R}{\mathbb{R}}

\renewcommand{\vt}[1]{\mathbf{#1}}
\newcommand{\surf}[1]{\mathcal{S}_{#1}}
\newcommand{\V}[1]{\mathcal{V}(#1)}
\newcommand{\I}{\mathcal{I}}
\renewcommand{\L}{\mathcal{L}}
\renewcommand{\P}{\mathcal{P}}
\newcommand{\dd}{\mathrm{d}}
\newcommand{\norm}[1]{\left\lVert#1\right\rVert}
\newcommand{\dotp}[2]{\langle#1\,,#2\rangle}
\newcommand{\dom}{\Omega}
\newcommand \eq[1]{\begin{equation}\begin{aligned}#1\end{aligned}\end{equation}}

\newcommand*{\inparagraph}[1]{\noindent\textbf{#1}\hspace{0.5em}}

\usepackage{pifont}% http://ctan.org/pkg/pifont

\usepackage{hyperref}
\usepackage{url}
\usepackage{booktabs}       % professional-quality tables
\usepackage{amsfonts}       % blackboard math symbols
\usepackage{nicefrac}       % compact symbols for 1/2, etc.
\usepackage{microtype}      % microtypography
\usepackage{caption}
\usepackage{makecell}
\usepackage{tikz,pgfplots}
\usepackage{tabularx}
\usepackage{multirow}
\usepackage{subcaption}
\usepackage{float}

\usepackage{graphicx}
\usepackage{amsmath}
\usepackage{amssymb}
\usepackage{amsthm}
\usepackage[nottoc]{tocbibind}
\usepackage{xspace}
\usepackage{wrapfig}

\usepackage[inline]{enumitem}

\usepackage[capitalize]{cleveref}
\crefname{section}{Sec.}{Secs.}
\Crefname{section}{Section}{Sections}
\Crefname{table}{Table}{Tables}
\crefname{table}{Tab.}{Tabs.}

\title{Implicit Neural Surface Deformation with \\Explicit Velocity Fields}

\author{Lu Sang$^{1,2}$, Zehranaz Canfes$^{1}$, Dongliang Cao$^{3}$,
Florian Bernard$^{3}$, Daniel Cremers$^{1,2}$ \\
$^{1}$Technical University of Munich, $^{2}$Munich Center of Machine Learning\\
\texttt{\{lu.sang, zehranaz.canfes, cremers\}@tum.de} \\
$^{3}$University of Bonn \\
\texttt{\{dcao, fb\}@uni-bonn.de} \\
}

\iclrfinalcopy % Uncomment for camera-ready version, but NOT for submission.
\begin{document}

\maketitle

\setlength{\intextsep}{0pt}

\begin{figure}[h]
    \centering
    \includegraphics[width=0.9\linewidth]{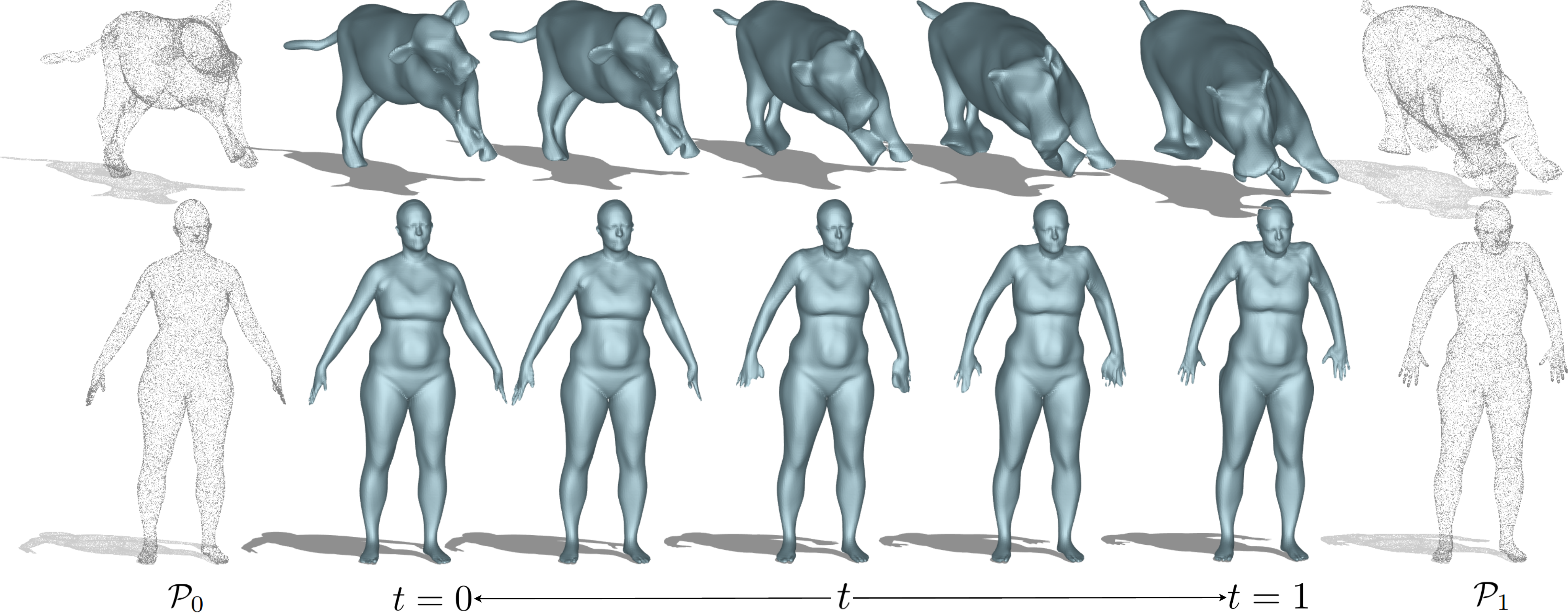}
    \caption{Given two point clouds $\P_0$ and $\P_1$, our method predicts a time-varying neural implicit surface that represents a smooth and physically plausible deformation from $\P_0$ to  $\P_1$. To ensure physical plausibility, we utilize a velocity network that leverages smoothness and divergence-free constraints.
    % to directly deform the time-varying implicit network. 
    % \sout{Given two input point clouds $\P_0$ and $\P_1$, our proposed method predicts a smooth time-varying neural implicit field \fb{does not make sense as writte; where does `timevarying' come from?}, which allows rendering physically plausible\fb{why are they physically plausible? that'd suggest an underlying physical model} intermediate shapes. \fb{both are correctly written in the abstract}}
    }
    \label{fig:enter-label}
\end{figure}

\begin{abstract}
  In this work, we introduce the first unsupervised method that simultaneously predicts time-varying neural implicit surfaces and deformations between pairs of point clouds.
We propose to model the point movement using an explicit velocity field and directly deform a time-varying implicit field using the modified level-set equation. This equation utilizes an iso-surface evolution with Eikonal constraints in a compact formulation, ensuring the integrity of the signed distance field. By applying a smooth, volume-preserving constraint to the velocity field, our method successfully recovers \textbf{physically plausible} intermediate shapes. % We employ a modified level-set equation to integrate neural implicit surface deformation with explicit velocity fields, eliminating the need for rendering explicit meshes. This equation utilizes iso-surface evolution with Eikonal constraints in a compact formulation, ensuring the integrity of the signed distance field. 
  Our method is able to handle both rigid and non-rigid deformations \textbf{without any intermediate shape supervision}. Our experimental results demonstrate that our method significantly outperforms existing works, delivering superior results in both quality and efficiency\footnote{the code is available: \url{https://github.com/Sangluisme/Implicit-surf-Deformation}}.
  % \todo{rewrite this.}

\end{abstract}

\section{Introduction}
Representing surfaces using implicit methods, such as signed distance fields, offers significant advantages over explicit methods in some applications.
For example, it allows flexible topological changes and is more memory-efficient compared to storing an explicit representation of a high-resolution surface.  Additionally, implicit representations allow for differentiable operations, as the respective surfaces are encoded in smooth fields, which in turn enhances a variety of downstream tasks, such as radiance field rendering by~\cite{yariv2021volume, neus2}.
Embedding a signed distance field within a neural network to represent a single surface demonstrated many successful outcomes, such as work from~\cite{sitzmann2019siren, gropp2020implicit, lars2018occupancy}. However, using implicit representations to model surface deformation or a dynamic surface evolution, especially with physically plausible deformations, still remains challenging. The challenges stem mainly from two inherent characteristics with implicit methods: 
\begin{enumerate*}[label=(\roman*),itemjoin=~]
\item implicit representations do not store explicit surface point locations, which makes it hard to directly manipulate surfaces during deformation.
\item the lack of traceable neighboring information in implicit fields prevents the use of efficient physical constraints, for example, as-rigid-as-possible regularisation, proposed by~\cite{sorkine2007rigid}, which is crucial in many mesh-based methods such as the work from~\cite{alexa2023rigid,eisenberger2021neuromorph,cao2024spectral}.
\end{enumerate*}
% \fb{use latex enum; preferably with (i) and (ii)}1) implicit representations do not store explicit surface point locations, which makes it hard to directly manipulate surfaces during deformation. 2) the lack of traceable neighboring information in implicit fields prevents the use of efficient physical constraints, such as for example the as-rigid-as-possible regularisation~\cite{sorkine2007rigid}, which is crucial in many mesh-based methods~\cite{alexa2023rigid,eisenberger2021neuromorph,cao2024spectral}. 
In this paper, we aim to tackle these core problems of implicit surface representations. To this end, we introduce a method that simultaneously recovers implicit neural representations of two given point cloud inputs, together with time-varying intermediate shapes between them. Most notably, our approach distinguishes itself from previous deformation methods based on implicit representations by recovering physically plausible intermediate shapes -- without supervision from ground truth intermediate shapes.
% Unlike most of the previous works can only create a linear interpolation between two given shapes~\cite{liu2022learning, Novello2023neural}.
To achieve this goal, we model the deformation of surface points by training a velocity network that utilizes smoothness and divergence-free constraints, thereby ensuring natural and physically plausible deformations. Our approach circumvents the need for mesh rendering during training, facilitating an end-to-end and fully differentiable training process. Our method supports both intrinsic and extrinsic deformations of the given point clouds, enhancing its versatility and application scope. In summary, we claim the following contributions:
\begin{itemize}
    % \item \sout{We propose a novel framework for recovering time-varying implicit representations network of shapes\fb{rephrase}, which allows for a physically plausible shape interpolation \fb{based on}}. 
    \item We propose a novel end-to-end framework that recovers the underlying surfaces of given point clouds together with physically plausible intermediate shapes.
    % \item We propose to model the deformation of the surface using velocity fields and apply suitable constraints to enable physically plausible surface deformation. 
    \item Our method directly deforms the implicit field by the explicit velocity field based on the level-set equation to avoid explicit mesh rendering.
    \item We propose to use a modified level-set equation that combines Eikonal constraint and thereby enables a compact joint optimization while preventing degenerated signed distance fields.  
    % \fb{mention that we propose a `novel modified level... eqn'}
    \item We validate our method on different datasets and demonstrate that our methods give rise to high-quality interpolations for challenging inputs, both quantitatively and qualitatively.
\end{itemize}

\section{Related Works}\label{sec:related_work}
\inparagraph{Surface representation methods} We roughly divide shape representation into explicit and implicit approaches. While explicit representations, such as polygon meshes, store mesh properties, e.g.\ vertices, edges, and faces explicitly, implicit methods encode the surface information into function fields, such as signed distance fields (SDF). 
With explicit methods, it is relatively straightforward to edit the shapes, since shape properties can directly be manipulated. However, there are some drawbacks to explicit surface representations. For instance, meshes can only have a fixed topology. It is not trivial to adapt vertices and the configuration of their connections (such as edges). 
% Further, meshes are resolution-dependent, which means that details are only preserved by high-resolution meshes while storing high-resolution meshes leads to large memory requirements. Besides, meshes are a discrete representation, so performing operations may not be differentiable.
% and when performing some operator on it, it is not differentiable. 
Implicit methods, on the contrary, allow arbitrary topological changes since no explicit surface and structural information are stored. Additionally, neural implicit representations enable arbitrary resolutions during inference, without memory increase during storage.
% with higher resolutions. 
% Respective methods encode the surface within a smooth field, facilitating differentiable operations. 
% Recently developed neural implicit methods, which encode surface information such as occupancies~\cite{lars2018occupancy, genova2019Deep} and signed/unsigned distance functions~\cite{park2019deepsdf, gropp2020implicit, chibane2020ndf} into neural networks, have demonstrated excellent performance in representing high-quality meshes.

\inparagraph{Mesh-based deformation} Mesh-based shape deformation is a well-studied problem in computer graphics. The most common strategy is to directly deform vertices based on some local deformation measurements (e.g.~as-rigid-as-possible (ARAP) proposed by~\cite{sorkine2007rigid}, PriMo~\cite{botsch2006primo}, etc.). Another direction is to deform intrinsic quantities like dihedral angles such as work from~\cite{alexa2023rigid,baek2015isometric} before reconstructing 3D shapes. Meanwhile, other works formulate shape deformation as a time-dependent velocity field~\cite{charpiat2007generalized,eckstein2007generalized} and incorporate specific constraints (e.g.\ volume preservation used by~\cite{eisenberger2018divergencefree,eisenberger2020hamiltonian}). Despite the great success achieved by mesh-based shape deformation methods, they rely on the local neighborhood information obtained from edges (or triangles) during deformation. Therefore, shapes have a fixed topology during deformation, which limits the applications for shapes with inconsistent topology or partiality. In contrast, our method directly works on an implicit surface, and thus has no constraint on the shape topology and is applicable for partial shapes based on spatial smoothness regularisation. In the experiment part, we demonstrate that our method is capable of deforming shapes with significant variations of shape resolution and partiality.  

\inparagraph{Implicit-field based deformation} Deforming implicit fields presents a significant challenge, as all information is encoded within a function field, preventing direct operations on the object. Previous works that studied this topic are typically for physical simulation, such as the work of~\cite{osher2003, museth2002, jones20063d}, and these works use classical discretely stored implicit fields. More recently, the widespread adoption of neural networks to represent implicit fields in shape modeling, work from\cite{yang2021geometry, sitzmann2019siren, ma2020neuralpull} inspired works to study shape deformation on implicit fields. Several works, e.g., \cite{peng2021animatable} utilizing neural networks for shape deformation have concentrated exclusively on human body movements , \cite{chen2021snarf, deng2019neural, bozic2021neuraldeformationgraphs} requiring prior information such as skinning details or intermediate point clouds . Others, like~\cite{cao2024motion2vecsets} have explored deforming implicit fields through generative or diffusion models, but these still necessitate intermediate point clouds for supervision. Some works have trained on datasets of shapes from specific object categories, \cite{deng2021deformed, genova2019learning, Iglesias_2017, hao2020dualsdf} aiming to deform from one category to another , rather than recovering plausible intermediate shapes. Some works address this problem by defining a latent space and getting a deformed shape via latent space interpolations~\cite{liu2022learning}.
\cite{yang2021geometry} introduced pioneering work that enables direct editing of implicit fields using user-defined handle points, ensuring the deformation remains consistent with the original object. Following this, \cite{mehta2022level} proposed using the level-set equation to deform the implicit network, providing theoretical insights into implicit field deformation. Building on this foundation, \cite{Novello2023neural} extended these concepts, applying them to 3D shapes, primarily focusing on smoothing surface deformations. In terms of directly deforming the implicit field using a velocity field, their work models the velocity field through linear interpolation between two pre-trained implicit networks representing the target shapes. Therefore, these works are limited to work on a pre-defined velocity field and fail to predict reasonable intermediate shapes when the deformation is not a linear translation. \par

% \fb{what is missing is a strong reason why the latter two works do not achieve what we do. a la `However, these works are limited in ..., which we address in this paper'}
In this paper, we adopt neural implicit surface representations and tackle the challenging problem of directly deforming the implicit field while recovering \textbf{physically plausible} intermediate shapes -- \textbf{without any rendering or intermediate ground truth supervision}. We achieve this by modeling deformations using a velocity field and directly deforming the implicit neural network using modified level-set-equations. Moreover, our training is end-to-end without any pre-trained SDF network needed.

\vspace{-0.3cm}
\section{Method}\label{sec:method}
\vspace{-0.3cm}
\setlength{\intextsep}{0pt}
\begin{wrapfigure}{r}{0.5\textwidth} 
% \vspace{-0.5cm}
    \centering
    \includegraphics[width=\linewidth]{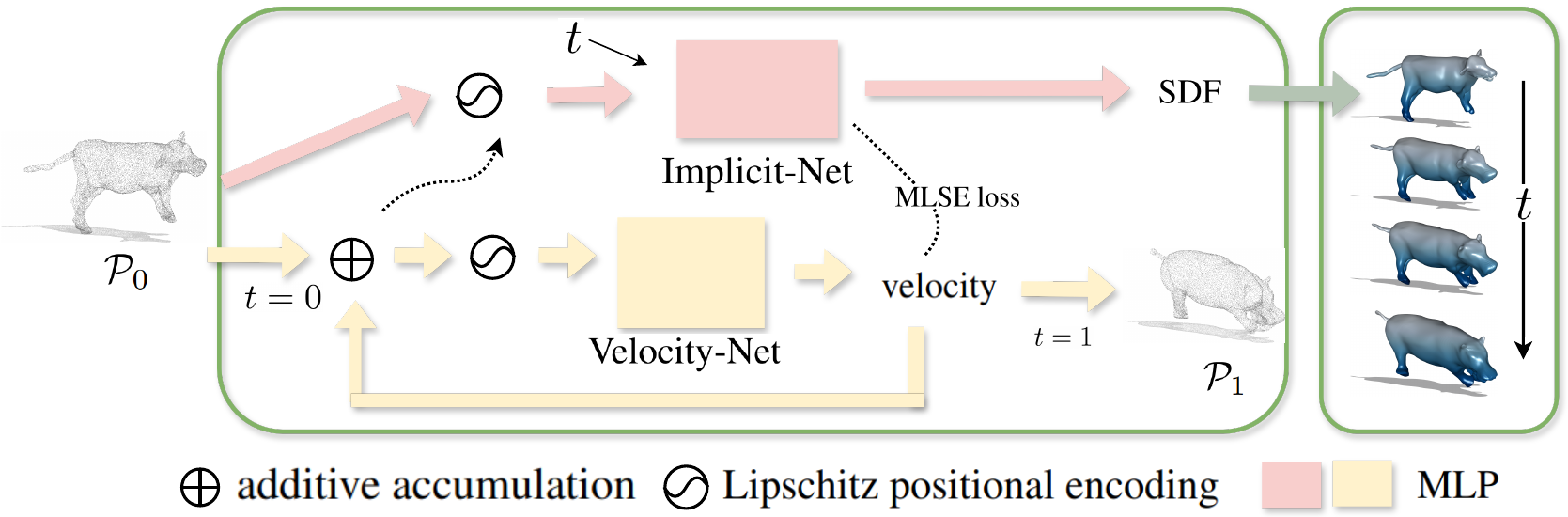}
    \caption{Pipeline of our method: given two point cloud $\P_0$ and $\P_1$, we train a time-varying Implicit-Net to predict SDF in different time steps and Velocity-Net to predict the velocity of the point at each time step. We directly deform the implicit field using MLSE loss.}
    \label{fig:pipeline}
\end{wrapfigure}
Given two 3D point clouds $\P_{0}=\{\vt{x}_i^0\}_i$ and $\P_{1}=\{\vt{x}_i^1\}_i$, we aim to reconstruct a time-varying implicit representation of the inputs together with natural and physically plausible intermediate surfaces. To this end,
% and their underlying intermediate shapes, 
we adopt a Lagrangian perspective from fluid mechanics to track the trajectory of individual points through a velocity field $\mathcal{V}:\dom \to \R^3$, for $\dom \subset \R^3$ being the point domain, and directly deform the time-varying implicit field $f: \dom \times \I \to \R$, where $\I = [0,1]$ is the time interval. 
% \sout{ \fb{first set the stage: what is our input, what is the desired output; afterwards talk about the reconstruction}.}
The primary challenges are 
\begin{enumerate*}[label=(\roman*), itemjoin=~]
\item modeling deformations that are realistic according to physical laws, i.e.\ modeling physically plausible movements, and \item deforming the implicit surfaces without relying on explicit mesh rendering or intermediate shape supervision.
\end{enumerate*}
% \fb{latex enum if possible with (i) and (ii)} 1) modeling deformations that are realistic according to physical laws, i.e.\ modeling physically plausible movements,
% \sout{\fb{we need to define what we mean by that, it can mean a lot of things that we don't mean}} 
% movements of surface points, 
% and 2) deforming the implicit surfaces without relying on explicit mesh rendering or intermediate shape supervision. 
To tackle these issues, we employ a smoothness constraint and a divergence-free constraint on the velocity field to ensure realistic movement. Additionally, we utilize a modified level-set equation to directly deform the time-varying implicit surfaces without rendering meshes explicitly. 
% We will provide detailed explanations of these methodologies in the subsequent sections.
% We train two neural networks, Implicit-Net, and Velocity-Net to model these two function fields.

\subsection{Time-Varying Implicit Fields}\label{subsec:time-sdf}
The time-varying implicit field $f(\cdot, \cdot)$ 
% \fb{$f$ is the field, $f(x,t)$ is not a field anymore, but a scalar; rewrite this,e.g. as $f(\cdot,\cdot)$ } 
takes a point location $\vt{x}\in\dom$ and a time $t\in\I$ as input, where $\I = [0,1]$ is the time interval and $\dom$ is the surface domain. Its purpose is to encode the evolution of a surface, that is, the shape $\surf{t}$ at time $t$ is represented by the zero-level-set of the implicit function $f(\cdot, t)$, i.e.  
\eq{\label{eq:time_implicit}
\surf{t} = \{ \vt{x} \in \dom | ~ f(\vt{x},t) = 0\}\;.
}
% Naturally, the given two point clouds also satisfy $\P_j \supset \surf{j} = \{ \vt{x} \in \dom | ~ f(\vt{x},t) = 0\}$, for $j=\{0,1\}$.

Signed-distance fields (SDFs) have many outstanding properties for representing surfaces: for $C^2$-smooth surfaces, the gradient of an SDF coincides with the surface normal direction $\vt{n}$ on the zero-crossing contour ($\partial \dom$),
% \fb{in the eq below there is nothing that states `zero crossing contour', instead, it is written for any $x$}
and the curvature $\kappa$ coincides with the divergence, that is 
\eq{\label{eq:normal_curvature}
\vt{n}(\vt{x},t) = \frac{\nabla f(\vt{x},t)}{\norm{\nabla f(\vt{x},t)}}\;, 
~~ \kappa(\vt{x},t) = \nabla \cdot \vt{n}(\vt{x},t), ~~ \text{for}~~\vt{x} \in \partial \dom\;.
}
% \todo{should I remove the $k$ term since we never used it?}
The time-varying signed distance function $f$ should also satisfy the Eikonal equation, since at every time $t$, $f$ still is a signed distance field, as proposed by~\cite{bothe2023mathematical}, i.e.\,
% \fb{why? explain in 2-6 words, or give a references} 
at any time $t\in\I$,
\eq{\label{eq:eikonal}
\norm{\nabla f(\vt{x},t)} = 1\;.
}
% To recover the time-varying signed distance field for representing surface deformations, we train an Implicit-Net, denoted as $f$, to approximate it.
% \fb{this sentence comes a bit out of nowhere; above we state mathematical properties, now suddenly we bvreak the reading flow. We should have some transition sentences that connects these}
\subsection{Velocity Fields}\label{subsec:velocity}
% To track the deforming that happens between the point clouds,
Inspired by the Lagrangian representation in fluid mechanics 
% \fb{above we say fluid mechanics}
, which tracks surface points by modeling the particle trajectory $\phi: \dom \times \I \to \dom$,
% \fb{not introduced; what is $\phi: .. \to ...$? }
% we propose to model the first-order derivatives of the particle trajectory, i.e.\ a velocity field, to track the points \fb{i don't get this sentence; rewrite}. 
we track the point by estimating its velocity and integrating the velocity field to form the point trajectory. 
The trajectory specifically consists of the position of particle $\vt{x}$ at time $t$. Assuming the points are moved by an external velocity field $\mathcal{V}:\dom \to \R^3$, where $\mathcal{V} \in V$, and $V$ is a Hilbert space of a smooth and compactly supported vector field on $\dom$. The velocity of the moving particle satisfies the following ordinary differential (ODE) equation with the initial condition: the trajectory derivative w.r.t. time $t$ is the velocity and the initial point location is given by $\vt{x}^0$
% the derivative of point trajectory $\phi$ \wrt time $t$ is the current point velocity $\mathcal{V}$, and the starting point position ($t=0$) is given as $\vt{x}^0$ 
% \fb{explain why/or what this ODE represents (and the init condition); again, verbally, 2-6 words}:
\eq{\label{eq:velocity}
\begin{cases}
& \V{\vt{x}} = \frac{\dd \phi(\vt{x}, t)}{\dd t}, ~~\text{for}~~ t \in \I \;, \\ 
& \phi(\vt{x},0) = \vt{x}^0 \;.
\end{cases}
}
Note that in our setting, we also enforce the ending point of particle trajectory by $\phi(\vt{x},1) = \vt{x}^1$, where $\vt{x}^1$ is the point in the target point cloud. To control the smoothness of the movement and ensure physical plausibility, we propose to constrain the velocity field by two aspects:  spatial smoothness of the velocity fields, and physical constraints.

\inparagraph{Velocity fields that generate diffeomorphisms} As the particle path $\phi: \dom\times t \to \dom$ represents a trajectory from one point cloud to another, we would like to recover a smooth transformation between two point clouds, which is consistent in both directions. Thus, we seek velocity fields that generate diffeomorphisms when integrated using~\cref{eq:velocity}, i.e.\ $\phi^{-1}(\cdot, t) = \phi(\cdot,1-t)$
% fb{negative time? the time interval is [0,1], so we could get -1} and 
\eq{\label{eq:velocity_field}
\phi(\vt{x}, t) = \vt{x}^0 + \int_0^t \V{\phi(\vt{x}, t)}\dd \tau\;.
}
% \todo{proof the existence of solution: Picard-Lindel\"{o}f theorem}
Inspired by~\cite{dupuis1998}, this can be achieved by regularizing on the space $V$ through the differentiable operator $\L = -\alpha \Delta + \gamma \vt{I}$ such that 
\eq{\label{eq:smoothness}
\norm{\mathcal{V}}_{V} = \norm{\L \mathcal{V}}_{l^2} = \int_{\dom}\norm{\L \V{\vt{x}}}_{l^2} \dd \vt{x} \;,
}
where $\vt{I}$ is the identity matrix. A more detailed explanation is provided in the~\cref{sec:appendix}.

\inparagraph{Divergence-free velocity fields} To model physically plausible deformations, we consider the basic conservation laws. One direct conservation law we can borrow is volume conservation. Since we move points on the surface along a trajectory, we assume that no particles are moved across the surface boundary at any time. Then, the total mass inside the surface stays the same, which directly follows from the divergence theorem~\cite{kreyszig2011}, i.e.\ 
\eq{\label{eq:divergence_free}
\nabla \cdot \V{\vt{x}} = 0.
}
A similar idea has also been explored in previous work from~\cite{eisenberger2018divergencefree, Cosmo2020} for the case of explicit polygon meshes. \par
% \eq{
% L^{\text{div}} = \int_{\dom} \norm{\nabla \cdot \V{\vt{x}}} \dd \vt{x} \;,
% }
\inparagraph{Velocity-Net integration} Our smooth velocity field is approximated by an MLP Velocity-Net $\mathcal{V}$. To track the point trajectory, we follow the forward Euler method for integrating ODEs, i.e.\, we take certain discrete time steps and integrate the velocity step by step using step length $\delta T = 1/ T$. Then, the discrete trajectory of points is formed by
\eq{\label{eq:trajectory}
\phi(\vt{x}, t+\delta t) = \phi(\vt{x}, t) + \V{\phi(\vt{x}, t)}\delta t\;.
}
The relation between $\P_0$ and $\P_1$ is then given as $\phi(\vt{x}, 0)=\vt{x}^0,~\vt{x}\in\P_0$ and $\phi(\vt{x}, 1)=\vt{x}^1,~\vt{x}\in\P_1$.

\subsection{Direct Implicit Field Deformation}\label{subsec:deformation}
In the previous sections, we introduced the time-varying implicit fields and velocity fields that represent the shapes and the deformation of points, respectively. In this section, we discuss how to directly deform the implicit field using the external velocity. We borrow the idea from fluid dynamics and treat every point as a fluid particle. Since points stay on the surface of any intermediate shape (i.e.~$f(\phi(\vt{x},t), t) = 0$ for any $t\in \I$ and $\vt{x}\in\partial \dom$)
% \fb{and $x \in ...$; i think we need to somehow state that this only holds for x on the surface}
, it implies there is no in- or outflow at the surface boundary  
% \fb{sentence does not make sense; what do we mean by `form as there is no ...'}
$\partial \dom$
\eq{\label{eq:lse1}
\frac{\dd}{\dd t} \int_{\dom} f(\phi(\vt{x},t),t) \dd \vt{x} = 0 \;.
}
Together with the initial condition, that is, the surface deforms from the underlying surface of point cloud $\P_0$, \cref{eq:lse1} implies that
\eq{\label{eq:lse2}
\begin{cases}
   & \partial _t f + \mathcal{V} \cdot \nabla f = 0 ~~ \text{in}~~\dom \times \I\;, \\
   & f(\vt{x}, 0) = f^0 \;,
\end{cases}
}
where $f^0(\vt{x})=0$ for $\vt{x}\in\P_0$. The linear transport in \cref{eq:lse2} is called the Level-Set Equation (LSE). However, the function $f$ is a signed distance function in our scenario, which means the Eikonal equation in~\cref{eq:eikonal} should also hold to prevent degenerated level-set functions. Previous work from~\cite{sussman1994level, sethian1996fast, sussman1999efficient} proposed to solve it by introducing a reinitialization equation at a pseudo time $\tau$, e.g.\, solving
\eq{\label{eq:reinitialization}
\frac{\partial }{\partial \tau} f + \text{sgn} (f^0)(\norm{\nabla f} -1), ~~ f|_{\tau =0} = f^0\;.
}
However, this requires solving an additional partial differential equation (PDE), and requires that the signed distance field $f$ at time $0$ is well initialized. To avoid pre-training a neural network to fit the starting mesh and solve the problem more compactly, we follow the idea of ~\cite{bothe2023mathematical} and~\cite{ fricke2023locally}, combining~\cref{eq:eikonal} and 
\eq{\label{eq:eikonal_grad}
\frac{\dd }{\dd t}\norm{\nabla f(\vt{x},t)} = -\norm{\nabla f}\dotp{(\nabla \mathcal{V})\frac{\nabla f}{\norm{\nabla f}}}{\frac{\nabla f}{\norm{\nabla f}}}  \equiv 0 \;,
}
with~\cref{eq:lse2} to get our \textbf{Modified Level-Set Equation} (MLSE) that reads 
% \fb{is this something new that we propose? if so, we should better emphasise this. `With that, we get our novel MLSE, that reads ...:'}\lu{equation are not new, they have it already.}
\eq{\label{eq:lse3}
\begin{cases}
   & \partial _t f + \mathcal{V} \cdot \nabla f = -\lambda_l f\mathcal{R}(\vt{x}, t) ~~ \text{in}~~\dom \times \I\;, \\
   & f(\vt{x}, 0) = f^0 \;,
\end{cases}
}
where $\mathcal{R}(\vt{x}, t) = - \dotp{(\nabla \mathcal{V})\frac{\nabla f}{\norm{\nabla f}}}{\frac{\nabla f}{\norm{\nabla f}}}$. Our MLSE in~\cref{eq:lse3} preserves the norm of the gradient at the zero-crossing contour. We also adapt the original proposed level-set equation in ~\cite{bothe2023mathematical, fricke2023locally} by adding a weight $\lambda_l$. We find that it helps to achieve better implicit surfaces while still preserving the desired properties. Compared to the original level-set equation~\cref{eq:lse2} and discretely enforced Eikonal constraint~\cref{eq:eikonal} on each time step, the modified level-set equation is more compact and solves a single partial differentiable equation (PDE) in an integrated way.\par 
% \fb{integrated? `in an integrated way' [or do we refer to computing integrals; if that's the case my statement does not make sense; neither does the original formulation, though]}. 
Our MLSE is the bridge between the velocity field and the implicit field.~\cref{eq:lse3} allows us to deform the implicit field without rendering explicit meshes. Moreover, every component of the formulation is differentiable, thus it enables end-to-end training. Our method jointly recovers the implicit surface for both given point clouds and intermediate shapes without the need for pre-training SDF neural implicit networks for given point clouds or meshes.
% By now, we have established a continuous constraint on both velocity fields, implicit fields, and the Eikonal equation. 
% \todo{add details in appendix.}
% Previous works~\cite{sussman1994} proposed to use reinitialization function $\partial_t f + \text{sng}(f^0)(\norm{\nabla f} -1) = 0$, where $\text{sgn}(f^0)$ is the sign of initial level-set function value. Dieter et. al point out that this 
\subsection{Loss}
We set up the training loss as follows. Velocity-Net loss $L_v$ contains smoothness (\cref{eq:smoothness}) and divergence-free (\cref{eq:divergence_free}) terms. Implicit-Net loss $L_f$ contains MLSE (\cref{eq:lse3} term.
% For the Velocity-Net $\mathcal{V}$ and Implicit-Net $f$, according to the analysis in~\cref{subsec:velocity} and \cref{eq:smoothness} and \cref{eq:divergence_free}, plus the modified level-set equation, the loss is \fb{restructure this sentence; it is a bit messy as it combines sections with eqns with MLSE}
\eq{\label{eq:velocity_loss}
L_v &= \int_{\dom} \norm{\L \mathcal{V}} \dd \vt{x} + \lambda_{\text{div}} \int_{\dom} |\nabla \cdot \mathcal{V} | \dd \vt{x} \;, \\
L_f &= \int_{\dom} |  \partial _t f + \mathcal{V} \cdot \nabla f + \lambda_l f\mathcal{R}| \dd \vt{x} \;.
} 
The divergence-free weight $\lambda_{\text{div}}$ can be set to $0$ to disable volume preservation.
% \eg, cross-category deformation. 
We show examples and the influence of the divergence-free term in the experiment section (\cref{sec:experiments}).

Finally, to fit the network to the given point clouds, we propose the matching loss 
\eq{\label{eq:matching}
L_m = \int_{\P_0} |f(\vt{x},0)| \dd \vt{x} + \int_{\P_1} |f(\vt{x},1)| \dd \vt{x} + \int_{\P_0^*} \norm{\phi(\vt{x},1) - \vt{x}^1} \dd \vt{x}\;.
}
The last term is used to indicate a double integral over $\dom$ and $\I$, which is needed as $\phi(\vt{x},1) = \vt{x}^0 + \int_0^1 \V{\phi(\vt{x}^0,\tau)} \dd \tau$. We use the forward Euler step, as described in~\cref{subsec:velocity} to integrate during training. Moreover, the last term also implies that one-to-one correspondence is needed for the given two point clouds. However, thanks to the spatial smoothness of the velocity field, we only need a small part of the given correspondence to achieve satisfactory results, thus $\P_0^* \subset \P_0$ is the set of points for which correspondence information is available. We will discuss the number of correspondences that are needed in~\cref{sec:experiments}. 
% Moreover,~\cref{eq:matching} means that one-to-one correspondence is needed for the given two point clouds. However, thanks to the spatial smoothness of the velocity field, we only need sparse correspondence to achieve satisfactory results, which we will show in the experiments section. 
% \todo{should we move this normal thing to the appendix?} Moreover, in case the normal information $\{\vt{n}\}^i$ is available for the given point cloud, the second-order constraints on matching can be added according to~\cref{eq:normal_curvature}, as
% \eq{\label{eq:normal_loss}
% L_n = \int_{\P_i} |1- \dotp{\frac{\nabla f}{\norm{\nabla f}}}{\vt{n}^i}| \dd \vt{x}, ~~\text{for}~~ i\in\{0,1\} \;.
% }
Our total loss term is defined as 
\begin{equation}
    L = \lambda_f L_f + \lambda_v L_v + \lambda_m L_m\;,
\end{equation}
% \fb{if space permits, we should make it a separate equation (not inline)}
where $\lambda_f$, $\lambda_v$, and $\lambda_m$ are weights to balance the joint training of velocity and implicit function term.
\section{Experiments}\label{sec:experiments}
\inparagraph{Neural network architectures and implementation} 
% As mentioned in the previous section~\cref{sec:method}, we need the neural networks to model a smooth and diffeomorphic transformation between shapes, 
% To ensure the smoothness and diffeomorphic transformation between shapes, as discussed in~\cref{sec:method}, for two neural networks, we choose the following architectures: (i) the Velocity-Net $\mathcal{V}$ outputs a velocity vector of the query point, it consists of 8-layer MLP with $256$ nodes for each layer; (ii) the Implicit-Net $f$ approximates the SDF of the query point at a given time, consists of $8$-layer MLP with $512$ nodes for each layer. For smoothness, we use Softplus~\cite{zhao2018} as an activation function. To utilize high-frequency information and maintain diffeomorphic (invertible), we incorporate Lipschitz continuous positional encoding~\cite{yang2021geometry} to model the complex signal. 
To ensure smooth and diffeomorphic transformations between shapes, as discussed in \cref{sec:method}, we use the following architectures for the two neural networks: (i) Velocity-Net $\mathcal{V}$ consists of 8-layer MLP with 256 nodes per layer; (ii) Implicit-Net $f$ also consists of 8-layer MLP with 512 nodes per layer. We use Softplus~\cite{zhao2018} as the activation function. To handle high-frequency information and maintain a diffeomorphism, we incorporate a Lipschitz continuous positional encoding~\cite{yang2021geometry}.
% However, directly incorporating the periodic functions leads to artifacts in the Implicit-Net $f$. 
% Instead, we use the Lipschitz continuous positional encoding~\cite{yang2021geometry}
% 

\inparagraph{Datasets} We evaluated our methods using several datasets: \textbf{Faust}~\cite{bogo2014faust}, \textbf{SMAL}~\cite{zuffi2017smal}, \textbf{SHREC16}~\cite{shrec16} and \textbf{DeformingThings4D}~\cite{li20214dcomplete}.
Faust and SMAL provide shapes with different categories and movements. Cross-category deformations involve non-rigid transformations between distinct objects, often with significant topological changes. Movement deformation involves changes in gestures within a single object, adhering to physical laws such as rigidity or conformality. DeformingThings4D provides continuous ground-truth displacements of the source mesh vertices in each frame. We used this data set to evaluate our interpolated meshes.
% The Faust and SMAL datasets provide shapes across different categories and movements. Cross-category (intrinsic) deformations involve transitioning between two distinct types of objects through non-rigid transformations. These deformations typically have significant topological changes. Movements (extrinsic) deformation is primarily represented by alterations in gestures within one single object. The deformations should adhere to physical laws, such as being rigid or conformal. 
% DeformingThing4D datasets provide continuous movements ground truth mesh of each object. Datasets contain the starting mesh and the displacements of each vertex at each time step. We render intermediate meshes using displacement. We utilize this dataset to assess the performance of our interpolated meshes. 

\inparagraph{Training strategy} To generate training data, we sample $20{,}000$ points on the surface of each mesh to create point clouds with partial correspondences. Each point cloud maintains ground-truth correspondences between $5\%$ to $20\%$ of its points 
% \fb{that is not sparse [i remember we call it `sparse' somewhere], that is up to 4000 correspondences!}. 
We jointly estimate the velocity and the time-varying signed distance field. To ensure the good initialized deformation of the implicit surfaces, we train $2{,}000$ warm-up epochs only for velocity fields. Then we gradually increase the loss term $L_f$ weight $\lambda_f$ for Implicit-Net. We implement our code using Jax~\cite{jax2018github} to enable fast higher-order derivative computations. We train for a total of $10{,}000$ epochs with batch size $4{,}000$. The run time is approximately $20$ minutes on a GeForce GTX TITAN X GPU with CUDA for each pair. 

% \inparagraph{Comparison methods} We compare our methods with both point cloud based implicit methods as well as mesh-based explicit methods. Point-cloud based: NFPG~\cite{yang2021geometry}, NIES~\cite{Novello2023neural}, LipMLP~\cite{liu2022learning}. Mesh-based: (LIMP~\cite{Cosmo2020}), Dongliang's work? 
\setlength\tabcolsep{0pt}
\setlength{\intextsep}{0pt}
\begin{figure}[b]
\vspace{-0.5cm}
    \centering
\includegraphics[width=0.9\linewidth]{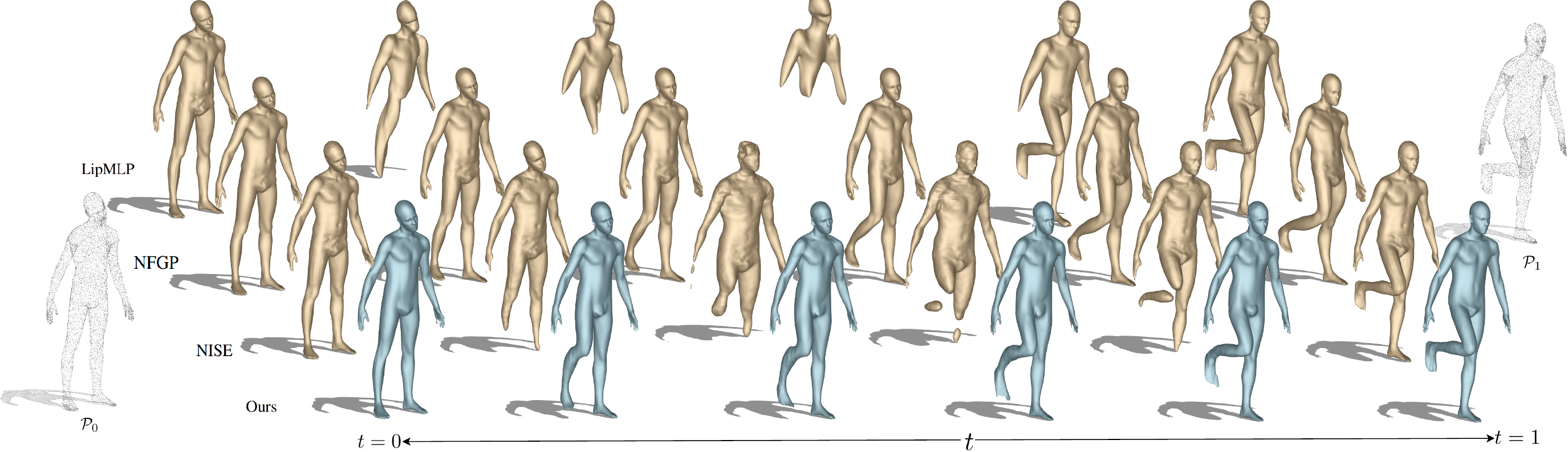}
   \caption{Experiment on extrinsic deformation. LipMLP~\cite{liu2022learning} and NISE~\cite{Novello2023neural} fail to estimate the physically plausible intermediate shapes. NFGP~\cite{yang2021geometry} can recover reasonable meshes but it is trained separately for each time step. Our method can recover realistic intermediate shapes in one model.}\label{fig:faust_5_4}
\end{figure}
\subsection{Shape Deformation}
We test our method on various deformation scenarios and compare it with other methods: LipMLP from~\cite{liu2022learning} uses MLP layers that satisfy Lipschitz continuity to ensure smooth transitions between source and target shapes. NFGP from~\cite{yang2021geometry} deforms shapes based on a source mesh and user-defined handle points. They estimate the implicit neural surface of the source mesh and then compute the deformed surface to match the target handle points. That means the method requires separate training for each time step. NISE fits two neural networks to the source and target meshes and trains an implicit field with a time dimension to estimate intermediate deformations. We use ground truth meshes to train both NISE and NFGP. NISE takes about 1.5 hours, and NFGP takes around 15 hours for each deformation step (over 75 hours for five steps). Our method requires only $1/5$ of the time compared to NISE, excluding the pre-training time for SDF networks of the source and target meshes, which takes around another 20 minutes.
% \fb{now I am curious how much it would be with the pretraining; we should provide this information here}.
% (note that in their code, they need to fit neural network to a mesh, that instead of giving the methods point clouds, we give the methods ground truth meshes for training).

% We set two Baseline methods: i) MLP only uses our Implicit-Net architectures and directly interpolates the time-space for recovering the intermediate shapes. ii) LipMLP~\cite{liu2022learning} uses MLP layers that satisfy Lipshitz continuities to provide continuous and smooth transactions between the starting and ending shapes.  We also compare with two other implicit representation-based methods NISE~\cite{Novello2023neural} and NFGP~\cite{yang2021geometry}. NISE~\cite{Novello2023neural} first need to fit two neural networks to the given point cloud and then train an implicit field with time-space (note that in their code, they need to fit neural network to a mesh, that instead of giving the methods point clouds, we give the methods ground truth meshes for training). 
% Moreover, it trains over 6 hours on one pair of shapes \todo{check if time is correct.} NFGP~\cite{yang2021geometry} just need starting shapes and some user-defined handle points to deform the shape, it only gives the final deformed shape, thus we use the corresponding points as handle points to deform the shape. For the intermediate shapes, we estimate the correspondence points' locations and give them to the algorithm as handle points. Notably, this method needs over 10 hours to deform a single step. \par

\inparagraph{Extrinsic (pose) deformation} Extrinsic deformation refers to only pose changes. 
% involves preserving local geodesic distances during the deformation process \fb{intuitively this sounds like intrinsic def., not extrinsic}, with
The transformations occurring within the same object category and no changes to the object type. In this context, we incorporate a divergence-free constraint (c.f.\ \cref{eq:divergence_free} with $\lambda_{\text{div}} > 0$) to ensure that only the object volume does not change during the deformation. \cref{fig:faust_5_4} illustrates the results on the Faust dataset, benchmarked against other methods. While other methods fail to produce physically plausible intermediate shapes, both NFGP~\cite{yang2021geometry} and our method successfully recover reasonable shapes. However, NFGP requires user-defined handle points for each step and must be trained incrementally, preventing it from generating a smooth deformed implicit neural surface. \par
% We trained it for $5$ middle steps and rendered the estimated mesh. 
\setlength\tabcolsep{0pt}
\setlength{\intextsep}{0pt}
\begin{figure}[t]
\vspace{-0.5cm}
    \centering
    \includegraphics[width=0.9\linewidth]{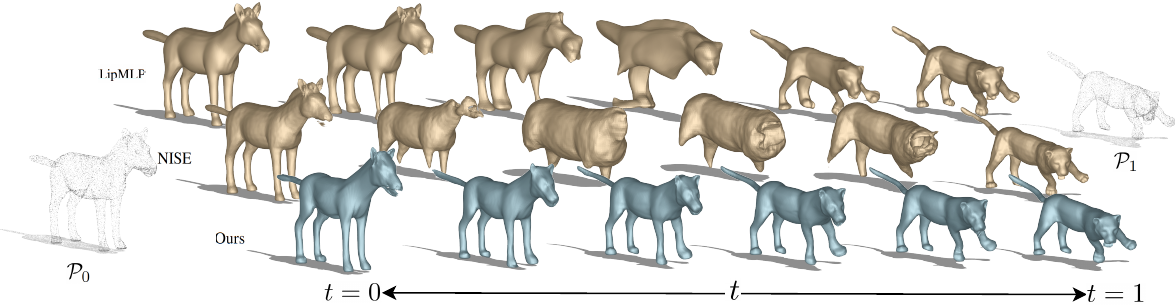}
   \caption{Experiment involves both extrinsic and intrinsic deformation. While LipMLP(~\cite{liu2022learning}) and NISE(~\cite{Novello2023neural}) fail to create reasonable middle-step meshes, our method generates appropriate transition meshes from two given point clouds.}\label{fig:smal_1_2000}
   \vspace{-0.5cm}
\end{figure}
\inparagraph{Instrinsic (non-rigid) deformation} Non-rigid deformation typically involves different objects for the source and target point clouds. In this case, we disable the divergence-free constraint by setting $\lambda_\text{div}=0$. \cref{fig:smal_1_2000} shows an example that includes different categories and poses deformation (intrinsic and extrinsic) in the source and target point clouds. Since NFGP~\cite{yang2021geometry} cannot handle non-rigid deformation, we only provide qualitative visualization results compared with the other two methods. While all methods can recover the source and target meshes, the comparison methods fail to generate reasonable intermediate meshes.\par
% The two baseline methods, MLP and LipMLP~\cite{liu2022learning} both directly fit a time-dependent implicit network at time $0$ and $1$. MLP is our Implicit-Net structure without Velocity-Net and LipMLP~\cite{liu2022learning} uses MLP layers that satisfy Lipshitz continuities. These two methods treat $t$ space as a latent descriptor and interpolate to get deformed intermediate shapes. However, the method cannot give physically plausible results in the middle
% NISE~\cite{Novello2023neural} first need to fit two neural networks to the given point cloud and then train an implicit field with time-space (note that in their code, they actually need to fit neural network to a mesh, that instead of giving the methods point clouds, we give the methods ground truth meshes for training). However, it fails due to its lack physically modeled velocity field. Moreover, it trains over 6 hours on one pair of shapes \todo{check if time is correct.} NFGP~\cite{yang2021geometry} just need starting shapes and some user-defined handle points to deform the shape, it only gives the final deformed shape, thus we use the corresponding points as handle points to deform the shape. For the intermediate shapes, we estimated the correspondence points' locations and gave them to the algorithm as handle points. Notably, this method needs over 10 hours to deform a single step. \par

\begin{figure}[ht!]
% \vspace{-0.8cm}
   \begin{minipage}{0.43\linewidth}
   \subcaptionbox{Average error evaluated on intermediate meshes. LipMLP~\cite{liu2022learning} and NISE~\cite{Novello2023neural} can fit well for the given meshes but produce big errors at the intermediate meshes. Our proposed method maintains a small error even on the middle shapes.\label{tab:comparison_error}}{\footnotesize
    \begin{tabular}{>{\centering\arraybackslash} m{1.7cm}|>{\centering\arraybackslash} m{1.2cm}|c|c}\toprule
    \multirow{2}{*}{\textbf{method}} & \multirow{2}{*}{\textbf{metric}} & \multicolumn{2}{c}{\textbf{datasets}} \\
    \cmidrule(l){3-4}
    & & fox & bear \\
    \midrule
     % \multirow{2}{*}{MLP} & ~~CD &  0.127 & 0.294\\
     % & ~~HD & 0.251 & 0.463\\
     % \midrule
     \multirow{2}{*}{\small{LipMLP}} & ~~CD$\downarrow$ &  1.745 & 2.649 \\
     & ~~HD$\downarrow$ & 2.456 & 1.401\\
     \midrule
      \multirow{2}{*}{\small{NISE}} & ~~CD$\downarrow$ & 0.178   & 0.366 \\
     & ~~HD$\downarrow$ & 0.262 & 0.560 \\
     \midrule
     % \multirow{2}{*}{\small{NFGP}} & ~~CD &   & \\
     % & ~~HD &  & \\
     % \midrule
     \multirow{2}{*}{Ours} & ~~CD$\downarrow$ &  \textbf{0.108} & \textbf{0.260} \\
     & ~~HD$\downarrow$ & ~~\textbf{0.114}~~ & ~~\textbf{0.265}~~ \\
     \bottomrule
\end{tabular}}
\end{minipage}%
\begin{minipage}{0.55\linewidth}
\subcaptionbox{Error plot of each intermediate mesh of bear dataset.\label{fig:error_graph}}{
    \input{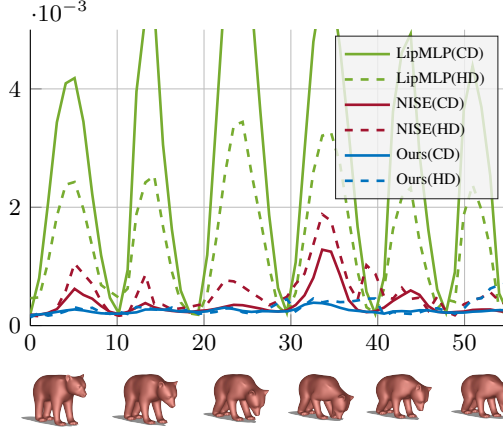}
   }
\end{minipage}%
    \caption{Quantitative evaluation of the deformed shapes. Chamfer Distance (CD) scaled by $10^3$ and Hausdorff Distance (HD) scaled by $10^2$ for the $55$ intermediate shapes.}
    \label{fig:error}
\end{figure}

\inparagraph{Quantitative evaluation} To quantitatively evaluate the interpolated meshes, we use the fox and bear animation from the DeformingThings4D~\cite{li20214dcomplete} dataset. These examples contain relatively small deformations per frame, making them suitable as ground truth baseline. Each sequence contains 55 meshes. For each dataset, we select 5 key meshes and estimate the deformations between them. We calculate the Chamfer Distance (CD) and Hausdorff Distance (HD) of the recovered meshes compared to the ground truth. We compare our method with LipMLP~\cite{liu2022learning} and NISE~\cite{Novello2023neural}. \cref{fig:error} shows the average error table over the 55 interpolated meshes (left) and the error plot for each mesh in the bear dataset (right). While all methods accurately recover the mesh at input time steps, LipMLP and NISE exhibit increasing errors at intermediate steps. Our method maintains a consistently low error on the intermediate meshes. Visualization results and error plots for the fox dataset are provided in~\cref{sec:appendix}.
\setlength\tabcolsep{0pt}

% \fb{fig 7 appearas above fig 6; they should have the right order}
\subsection{Incomplete and Sparse Input}
Implicit methods showcase remarkable flexibility in representing shapes with varying typologies. In this section, we show some challenging cases that our method can still tackle.\par
\begin{wrapfigure}{r}{0.6\textwidth} 
% \vspace{0.5cm}
    \includegraphics[width=0.9\linewidth]{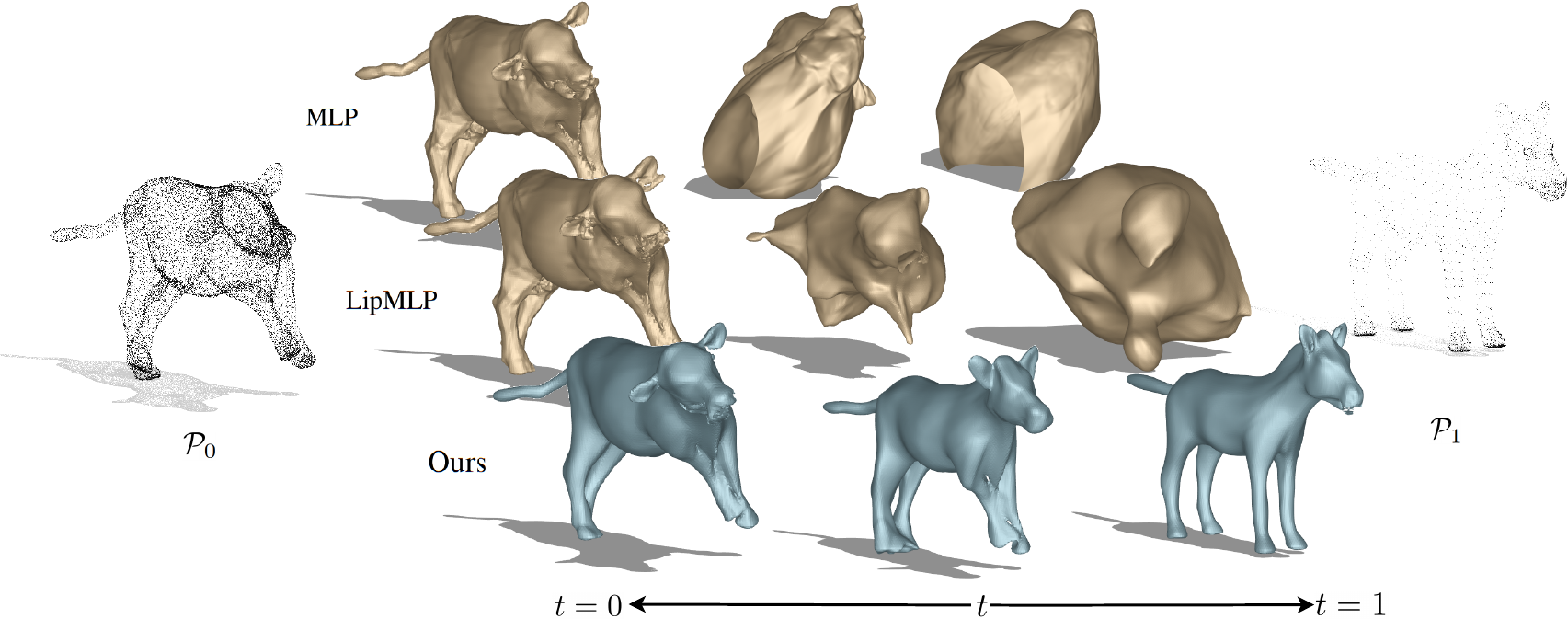}
    \caption{Compared to previous methods, our method can still recover both the target shape and the intermediate shapes even one input point cloud is excessively sparse.}\label{fig:sparsity}
\end{wrapfigure}

\inparagraph{Different sparsity inputs} Most existing approaches require fitting two separate networks to estimate the Signed Distance Fields (SDF) for the initial and final shapes, as highlighted in previous works~\cite{Novello2023neural, yang2021geometry, liu2022learning}. Consequently, the quality of the results is heavily dependent on the characteristics—such as the sparsity of the source and target point clouds. If the fitting process for one shape is unsuccessful, these methods fail to estimate intermediate shapes. Our approach overcomes these challenges through the Velocity-Net, which enables tracking the dense initial point cloud $\P_0$ to the sparse final point cloud $\P_1$, and recovering the underlying shapes without compromising result quality.~\cref{fig:sparsity} illustrates the varying sparsity levels of the input data. The source point cloud $\P_0$ contains $\sim 20{,}000$, points while the target point cloud $\P_1$ only has $\sim 2{,}000$ points. While the source point cloud is dense enough to train a neural network for fitting an SDF, the target point cloud is much sparser than the source. Although one could employ densification strategies~\cite{sang2023weight, zhao2022selfsupervised} or utilize priors \cite{ma2022reconstructing} before fitting the networks, our method successfully reconstructs both the final and intermediate shapes without any additional modifications.\par
\begin{figure}[t]
\vspace{-0.5cm}
    \centering
    \includegraphics[width=0.9\linewidth]{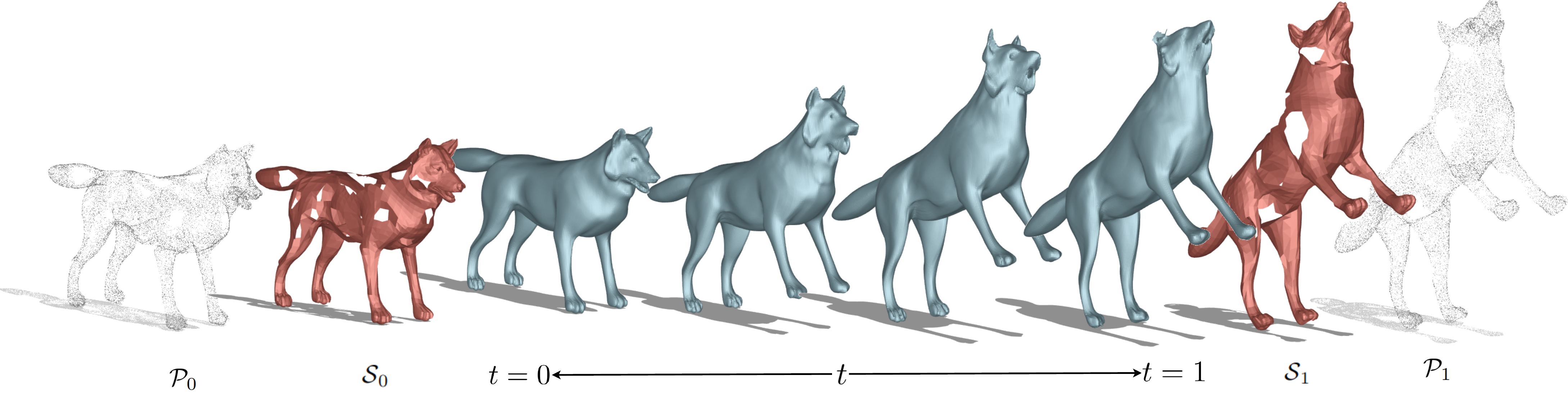}
    \caption{We sample point cloud $\P_0$ and $\P_1$ from ground truth meshes with holes $\surf{0}$ and $\surf{1}$, respectively. The input point clouds are incomplete. The proposed method can still recover neural implicit surfaces with complete shapes together with the intermediate steps.}
    \label{fig:incomplete_shape}
    \vspace{-0.5cm}
\end{figure}

\inparagraph{Incomplete inputs} Additionally, our time-varying implicit network is capable of completing the shape even when both input sets are incomplete. We study the situation that the given point clouds are incomplete in different areas. We sample point clouds from the incomplete shapes $\surf{0}$ and $\surf{1}$ to create point clouds with holes $\P_0$ and $\P_1$ as input data. Results in~\cref{fig:incomplete_shape} demonstrate that our method can complete the shape. This experiment result implies that we do not need consistent topology from the inputs, i.e.\, we can handle inputs with different topology features. Note that this is typically challenging for mesh-based methods, both with and without ground truth correspondences. Meshes have a fixed topology and deforming a mesh to another one with substantially varying topology is a challenging endeavour.
% it is hard to deform a mesh to another whose vertices and faces are different. 
Moreover, mesh-based methods struggle with completing the mesh without giving a complete shape as a prior. We provide an analysis and comparison with mesh-based methods in~\cref{sec:appendix}.

\subsection{Ablations}
\begin{figure}[t]
    \centering
    \includegraphics[width=0.9\linewidth]{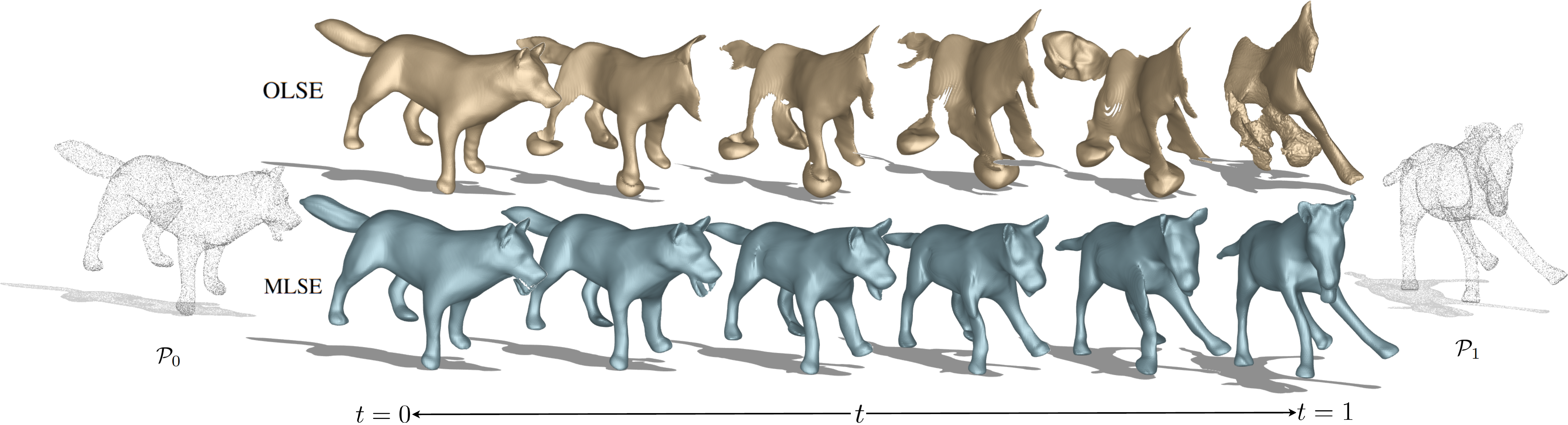}
     \caption{Ablation study for original level-set equation (OLSE)~\cref{eq:lse2} (first row) and modified level set equation (MLSE)~\cref{eq:lse3} (second row). The OLSE does not coincide with the eikonal constraint~\cref{eq:eikonal} while MLSE implies it. The reconstruction results show the MLSE produces a more stable and non-degenerate signed distance field.}\label{fig:smal_1037_2037}
     \vspace{-0.7cm}
\end{figure}

\inparagraph{Modified level-set equation}
In this section, we demonstrate that the modified level-set equation~\cref{eq:lse3} leads to more stable results compared to the original level-set equation. 
% Novello \etal~\cite{Novello2023neural} and Mehta \etal~\cite{mehta2022level} adopt the original level-set equation~\cref{eq:lse2}, however, instead of modeling an explicit velocity fields, their works are more theoretical analysis and only use interpolated velocity fields. Moreover, they do not use continuous Eikonal equation regularization. 
% We show that the modified level-set equation helps in particular for deformation across category \todo{anti-isometic?} and has large pose deformation at the same time. 
\cref{fig:smal_1037_2037} shows the qualitative results of the two different level-set equations. The Original Level-Set Equation (OLSE) uses the formulation~\cref{eq:lse2} plus~\cref{eq:eikonal} at every discrete time step in training. 
% When the deformation is non-rigid and the extrinsic distortion is large, it results in degenerated signed distance fields. However, t
The Modified Level-Set Equation (MLSE) embeds the Eikonal constraint compactly.~\cref{fig:smal_1037_2037} shows that MLSE prevents the degenerated meshes.
% ; instead, it enforces the constraint in one term \fb{not one term; this sounds weird; rephrase}and generates stable signed distance fields.
\setlength\tabcolsep{0pt}

\inparagraph{Volume preservation effect}
We explore the influence of our divergence-free regularizer~\cref{eq:divergence_free} proposed in~\cref{subsec:velocity}. We show that the divergence-free regularizer on the velocity field indeed preserves the total volume of the recovered shape.~\cref{fig:div_ablation} shows that if the divergence-free constraint is enforced, the network produces a surface that preserves the volume of the source point cloud. It generates meshes that adopt the appearance features of the target point cloud but do not expand the volume to fit the target point cloud. The whole visualization is provided in~\cref{sec:appendix}.
\setlength\tabcolsep{0pt}
\newcolumntype{C}{ >{\centering\arraybackslash} m{0.9cm} }
\begin{figure}[ht!]
\vspace{0.2cm}
    \centering
    \begin{minipage}{0.5\linewidth}
    \subcaptionbox{Visualization of the output meshes with and without divergence-free loss. From the same source point cloud, the left side of the mesh is slim compared to the target point cloud while the right side mesh can fit perfectly.  
    \label{fig:div_ablation}}{
         \includegraphics[width=0.9\linewidth]{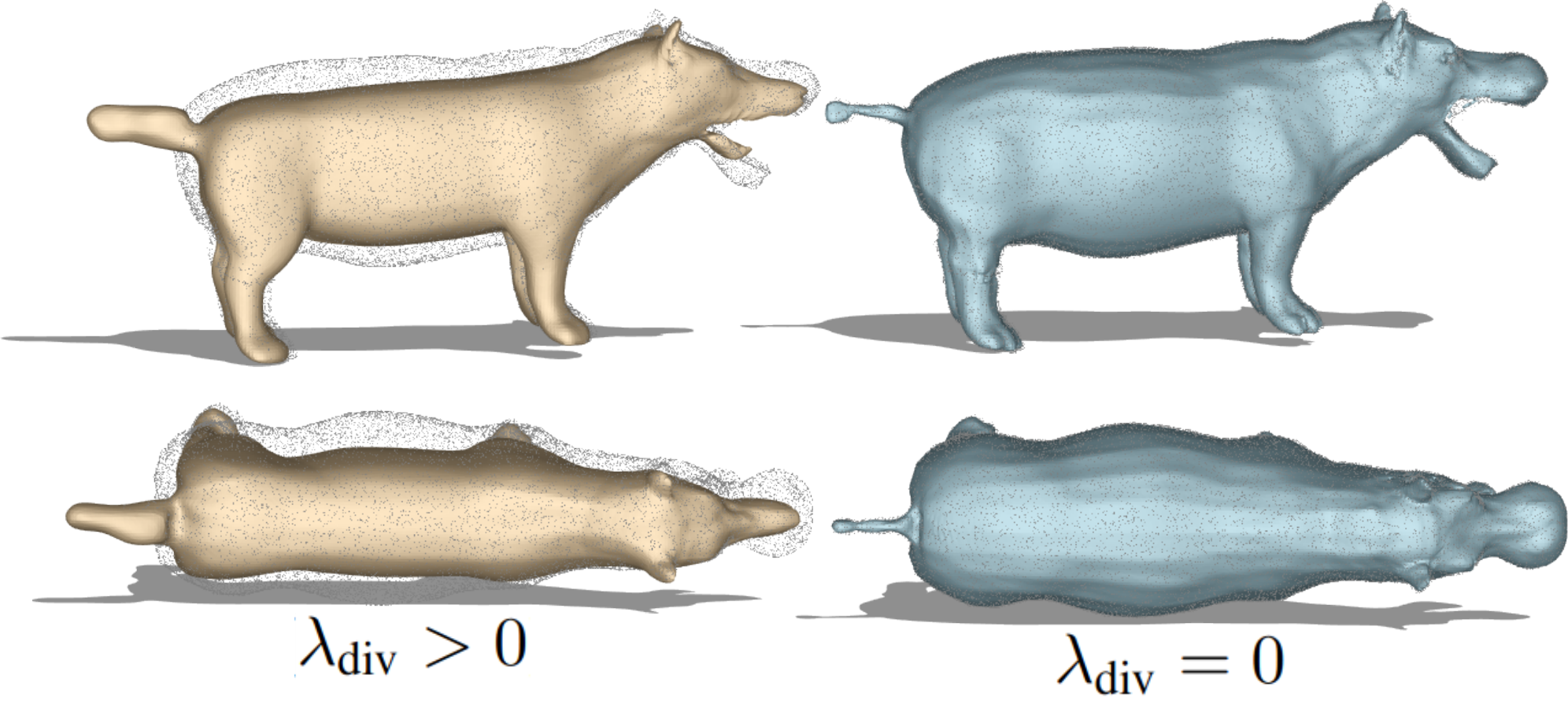}}
    \end{minipage}%
     \begin{minipage}{0.48\linewidth}
          \centering
          \subcaptionbox{Error with different sparsity levels. The reconstruction accuracy increases along with the number of GT correspondences.
          % Both errors decrease along the number of ground truth vertices\fb{this sentence does not make sense}. 
          However, even with $1\%$ ground truth correspondence, our method can get small CD and HD errors.\label{fig:sparsity_level}}{
    \begin{tabular}{ >{\centering\arraybackslash}m{1.4cm}| >{\centering\arraybackslash} m{1cm}|C|C|C|C}\toprule
    \textbf{dataset} & \textbf{Metric} & $1\%$ & $5\%$ & $10\%$ & $20\%$ \\
    \midrule
    \multirow{2}{*}{fox} & CD$\downarrow$ & 0.119 & 0.108 & 0.111 & 0.110  \\
     \cmidrule{2-6}
    & HD$\downarrow$ & 0.148 & 0.141 & 0.114 & 0.106 \\
    \midrule
    \multirow{2}{*}{bear} & CD$\downarrow$ & 0.257  &  0.250 &  0.251 & 0.250 \\
    \cmidrule{2-6}
    & HD$\downarrow$ &  0.310 & 0.306 & 0.308 & 0.291 \\
     \bottomrule
    \end{tabular}
    }
    \end{minipage}%
    \caption{Divergence-free constrains ablation (left) and Sparsity correspondence ablation (right).}
    % \vspace{-0.5cm}
\end{figure}

% \inparagraph{}
% Compared to mesh-based method, one big advantage is the that implicit method makes no-assumption on topology of the input. It can work with shapes with abitrary complex topological features include self-intersections or missing structures. The reasons are typically implicit methods work on point-cloud which does not have neighbouring connection or constraint like water-tight. However, one challenges of implicit neural fitting is that the input point cloud is too sparse. A typical solution would be different densification strategies~\cite{sang2023weight, zhao2022selfsupervised} or using priors~\cite{ma2022reconstructing}. Our proposed paper shows supreme performance when encounter inconsistent-topology during shape interpolation and provide another way to bypass the sparity point cloud problems and recover high-quality and more completed surfaces.  

% \inparagraph{Lipschitz positional encoding}

\inparagraph{Sparsity of the correspondences}
Our method utilizes a certain amount of ground truth correspondence. In this section, we explore the influence of correspondence numbers on the deformation qualities.  We sample correspondences in different proportions to the point cloud numbers: $1\%$, $5\%$, $10\%$, $20\%$ to test the recovered intermediate mesh quality.
% \fb{no result presented...; refer to figure}
% \input{figure_tex/sparsity_ablation}

\begin{figure}[ht]
    \centering
    \includegraphics[width=0.9\linewidth]{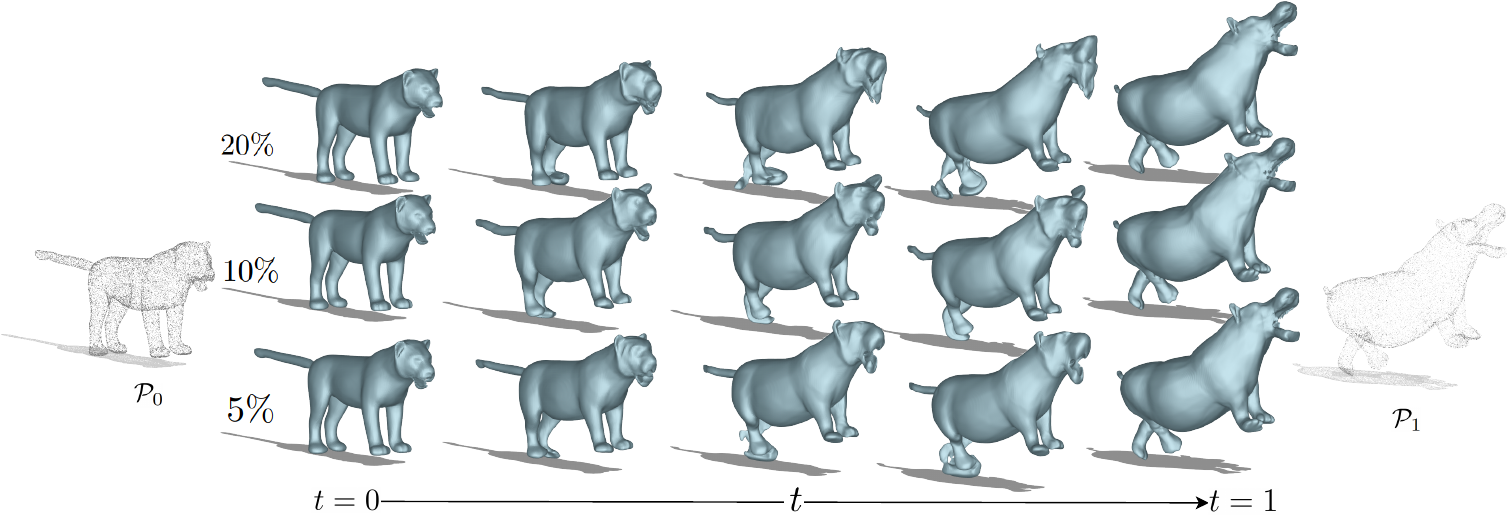}
    \caption{ We add noise to the correspondences by randomly choosing different portions of the GT correspondences and swap them with its $5th$ nearest neighbor correspondences. Qualitative results show that our method is stable up to $10\%$ correspondences and still gives relatively reasonable results up to $20\%$ misaligned correspondences.
    }
    \label{fig:nosiy_ablation2}
\end{figure}

\inparagraph{Noisy correspondences ablation}
We test the robustness of our method in different ways. First, we test against local noise on ground-truth correspondences. The test data contains $5\%$ correspondences with respect to the total number of input points. We sample $5\%$, $10\%$, and $20\%$ of the correspondences and swap them with their $5$th nearest neighbor correspondences (see~\cref{fig:nosiy_ablation2}). 
%Note that we swap correspondences rather than points, and considering the sparseness of the correspondences, they are located at a valid distance. Our method still produces satisfactory results with $20\%$ of correspondences being misaligned. 
We also test against the global noise, where we randomly swap them with other correspondences, regardless of whether the swapped correspondences are neighboring. This represents an extreme case for noise simulation. Due to page limitation, we only show the local cases in the main paper. For more ablations please refer to the appendix. 

\begin{figure}[ht]
    \centering
    \includegraphics[width=0.8\linewidth]{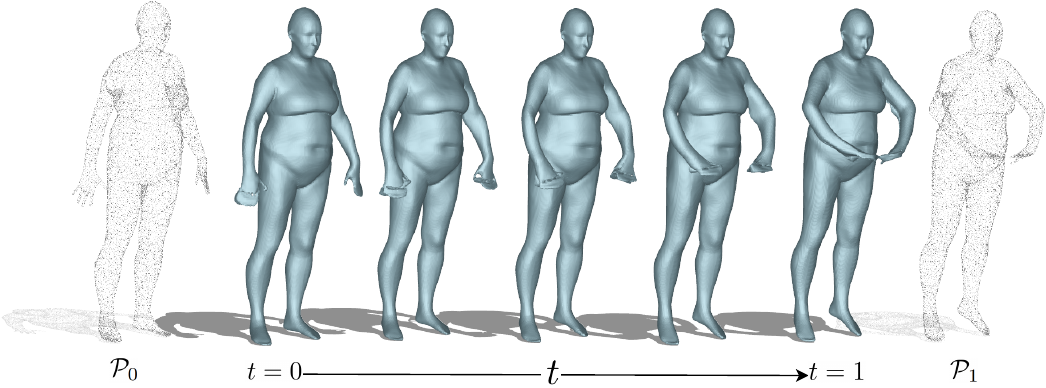}
    \caption{Results using correspondences obtained by other methods~\cite{cao2024spectral}. The method offers $5000$ correspondences to each pair of the shapes.}
    \label{fig:faust_cao}
    \vspace{-3mm}
\end{figure}

\inparagraph{Combining other methods to obtain correspondences}
Our proposed method integrates seamlessly with existing point-registration techniques when ground-truth correspondences are unavailable. In this section, we present results using a prior method~\cite{cao2024spectral} to first obtain correspondence vertices, followed by the application of our method. Many point-registration approaches provide only sparse correspondence pairs with small deviations. However, due to the robustness of our method and the fact that we require only around $10\%$ of correspondences to achieve strong results, we can effectively utilize most of their output. \cref{fig:faust_cao} illustrates the results of our method built on top of a matching technique. For more results using obtained correspondences, please refer to~\cref{sec:appendix}.
\section{Discussion}\label{sec:conclusion}
\inparagraph{Limitations and future works} Even though our work is self-supervised, we still need sparse correspondences of the point clouds. Moreover, due to the lack of neighboring information, our method struggles with large deformations and may introduce artifacts around the reconstructed surfaces (c.f.\ ~\cref{sec:appendix}~\cref{fig:failure_cases}). In the future, we will explore dealing with large deformations and extend our work to dynamic implicit surface generation.\par 
% We will eliminate the need for ground truth correspondences in future work 
% \fb{this sounds like it is trivial and raises the question why we did not alreaedy do it. Rather argue that doing so is an itneresting direction for future work, where particular challenges lie in x, y, z.}and adopt more local constraints on the velocity field to improve the ability to estimate large deformation.\par
\inparagraph{Conclusion} 
In this paper, we introduce a method to recover the implicit surface of two given point clouds based on sparse correspondences, while also generating a natural and physically plausible intermediate deformation. Our approach does not require any intermediate shape supervision beyond the provided source and target point clouds.
% Moreover, we can recover the physically plausible intermediate shapes.
% Shapes and deformations are represented as a time-varying implicit neural network that enables rendering shapes at arbitrary time and resolution.  Our method explores the direct deformation of the implicit field using an explicit estimated velocity field, which allows us to estimate the deformation directly from the point cloud input, enables us to represent more flexible topologies, and deals with more challenging scenarios. 
 Our method directly deforms the implicit field using an explicitly estimated velocity field, enabling us to estimate deformations directly from the point cloud input. This approach allows for the representation of more flexible topologies and can handle more challenging scenarios. Our method also broadens the application of implicit representations from static objects to dynamic objects.\par

% \fb{I suggest to restructure this last section: separate limitations and future work into another section (above conclusion). in the conclusion we have the summary (as is), and additionally we include 2-4 sentences of the big picture: why is our method important, what impact on adjacent and applied topics do we see? why is our work relevant, etc}
% \sout{the smoothness constraint is still too weak for very large pose deformation \fb{what do we mean by that? can show a failure case (eg in supp mat)}. Some artifacts might be introduced around the desired surface and the movement is unsatisfying, see analysis in}~\cref{sec:appendix}. 

\newpage

\bibliography{iclr2025_conference}
\bibliographystyle{iclr2025_conference}

\newpage 

\appendix

\section{Appendix}\label{sec:appendix}

\subsection{Mathematical Proofs}

\paragraph{Existence of the velocity field} We assume that our velocity field is the solution of~\cref{eq:velocity}. The existence of the solution for the differential equation is given by Picard-Lindel\"{o}f Theorem~\cite{arnold1992ordinary} which states that for $\phi:\R\times\R^n \to \R^n$, for continuous $t$ and Lipschitz continuous $\mathcal{V}$, then there exist an $\epsilon >0$ such that the initial value problem~\cref{eq:velocity} have unique solution $\mathcal{V}$ on interval $[t-\epsilon, t+\epsilon]$. As we choose $\mathcal{V}$ smooth enough, the requirement is satisfied. 

\paragraph{Velocity fields that generate diffeomorphism} 
\cite{dupuis1998} proved the existence of the smooth trajectory generated by~\cref{eq:velocity} depends on the smoothness constraint in the vector field $V$. They also proved that choosing $V$ such that $V$ is a smooth and compactly-supported vector field with an inner product defined by a differential operator $\mathcal{L}$ ensures the solution in the space of diffeomorphism. 

\paragraph{Smooth operator $\mathcal{L}$} The differentiable operator $\mathcal{L}$ introduced in~\cref{eq:smoothness} is chosen to have the type $ \mathcal{L}= -\alpha \Delta + \gamma \vt{I}$, where $\alpha$ enforces the smoothness and $\gamma > 0$ ensures the operator is non-singular. In the experiments, we set $\alpha = 0.01$ and $\gamma = 1$. The velocity field is a Hilbert space defined by the operator $\mathcal{L}$ with norm
\eq{
\norm{\mathcal{V}}_V = \dotp{\mathcal{V}}{\mathcal{L}\mathcal{V}}\;.
}
~\cite{beg2005} proved that this type of choice for operator $\mathcal{L}$ stratifies the requirement that $\phi$ is a diffeomorphism~\cite{dupuis1998variational}.

\paragraph{Level-set equation} Following~\cref{eq:lse1} and smoothness assumption of $\dom$ and $f$ we have
\eq{
\int_{\dom} \frac{\dd f(\phi(\vt{x},t)}{\dd t} \dd \vt{x} = 0\;,
}
we have $\frac{\dd f(\phi(\vt{x},t)}{\dd t} = 0$ almost everywhere. Compute the derivatives, we have
\eq{
\frac{\dd f(\phi(\vt{x},t)}{\dd t} = \partial f_t + \partial f_{\vt{x}} \partial \phi_t = 0\;.
}
Together with~\cref{eq:velocity} and $\partial_{\vt{x}} f = \nabla f$, we have 
\eq{
\frac{\dd f(\phi(\vt{x},t)}{\dd t} = \partial f_t + \nabla f \cdot \mathcal{V} = 0\;,
}
which is the original level-set equation.\par
\inparagraph{Modified Level-set equation} To ensure the Eikonal constraint on continuous time-space for any $t$, it is equivalent to solving an additional initial problem of the PDE, i.e.\,
\eq{\label{eq:eikonal2} 
\begin{cases}
& \frac{\dd}{\dd t}\norm{\nabla f(\vt{x}, t)} = 0, ~~ t\ \in \I\;, \\
& \norm{\nabla f(\vt{x}, 0)} = 1\,.
\end{cases}
}
The function above ensures that $\norm{\nabla f(\vt{x},t)} = 1$ for any $t\in \I$.
\cref{eq:lse3} holds on the zero-crossing surface $\partial \dom$ because the function value $f$ is zero at the zero-cross surface and the two term $\partial f_t +\mathcal{V} \cdot \nabla f$ and $\mathcal{R}$ both equal zero in the surface domain $\dom$. It is more compact to solve~\cref{eq:lse3} than solve~\cref{eq:lse2} plus~\cref{eq:eikonal2} separately since the later solves two PDEs~\cite{bogo2014faust}. \par
Note that~\cref{eq:lse3} needs two initial value $f(\vt{x}, 0) = f^0$ and $\norm{\nabla f(\vt{x}, 0)} = 1$. We explain in the~\cref{subsec:training_strategy} how we avoid pre-training a network to satisfy the initial condition.

% \todo{maybe move this to the appendix? since this is not our contribution?}
% diffeomorphism $\phi(\cdot, t)$ under the condition that $\mathcal{L}$ is smooth enough.

% follow the existence of trajectory that generated by~\cref{eq:velocity} in ~\cite{dupuis1998variational}, we explain the smooth operator $\mathcal{L} = -\alpha \Delta + \vt{I}$.
% For Hilbert space $V$, the inner-product is defined  through a differential operator $L$, such that for any function $u, v \in V$, the inner product is defined by
% \eq{
% \dotp{u}{v}_{V} = \dotp{Lu}{Lv}_{l^2} = \dotp{L^*Lu}{v}_{l^2} \;,
% }
% where $\dotp{\cdot}{\cdot}_{l^2}$ is usual $l^2$ product in Euclidean space and $L^*$ is the self-adjoint operator of $L$. 
\subsection{Training Strategy} \label{subsec:training_strategy}
We implement our method using Jax~\cite{jax2018github} and set the learning rate to $0.005$ with a decay rate $0.5$ within interval $2000$. We initialize Implicit-Net's weights and bias such that it represents a sphere at step $0$, following the method proposed in~\cite{gropp2020implicit}.\par
\inparagraph{Invertible Lipschitz Positional Encoding}
We adopt the invertible Lipschitz positional encoding same as~\cite{yang2021geometry} to cooperate with MLP in both velocity net and implicit net to produce a stable output.
\eq{\label{eq:positional_encoding}
\gamma_i(\vt{x}) = \frac{1}{\sqrt{2m+1}}(x_i, \frac{\cos(2^0\pi x_i)}{2^0\pi}, \frac{\sin(2^0\pi x_i)}{2^0\pi}, \dots, \frac{\cos(2^m\pi x_i)}{2^m\pi}, \frac{\sin(2^m\pi x_i)}{2^m\pi} ) \;.
}

\inparagraph{Oriented point cloud} Our method does not require an oriented point cloud (point cloud with normal). However, if the normal information $\{\vt{n}\}^i$, for $i=\{0,1\}$ is available for the given point clouds, the second-order constraints on matching loss can be added according to~\cref{eq:normal_curvature}. The normal loss term is
\eq{\label{eq:normal_loss}
L_n = \int_{\P_i} |1- \dotp{\frac{\nabla f}{\norm{\nabla f}}}{\vt{n}^i}| \dd \vt{x}, ~~\text{for}~~ i\in\{0,1\} \;.
}
$L_n$ term can accelerate the convergence speed of Implicit-Net at time $t=0$ and $1$. \par

\inparagraph{Network initialization} We initialize Implicit-Net such that it represents a unit sphere at time $0$~\cite{gropp2020implicit}. Thus it is a valid signed distance field that satisfies the initial condition in~\cref{eq:eikonal2}. To satisfy the initial condition in~\cref{eq:lse3} without pre-train the net on $f(\vt{x},0)$, we set $\lambda_m$ much larger than $\lambda_f$ such that the network first converges at time $0$ to fit $f(\vt{x},0) =0$ on the given input point cloud $\P_0$. Thus, for the experiment showed on the paper, we set $\lambda_f = 100$, $\lambda_m = 200$, $\lambda_v = 20$ and $\lambda_l = 10$.

\inparagraph{Warm up training} As described in~\cref{sec:experiments}, we first freeze the implicit network, i.e.\, we set $\lambda_f = 0$, $\lambda_v = 20$, $\lambda_m = 100$ for first $2000$ epochs, then we gradually increase the it using $\lambda_f = \frac{k-2000}{n-2000}100$ for $k<n$, and $\lambda_f = 100$ for $k\ge n$, where  $k$ is the $k$th-epoch, and $n=5000$. As we observe that velocity field convergence is faster than the implicit field, we decrease the velocity loss to train only implicit net after a certain epoch, i.e.\ $\lambda_v = 0$ for $k>8000$. 

\subsection{Quantitative Results}
\begin{figure}
\centering
\begin{minipage}{0.55\linewidth}
\subcaptionbox{Visualization of the middle step meshes for fox datasets. We visualize the deformed meshes for middle step $5$, $15$, $25$, $35$, and $45$ of NISE~\cite{Novello2023neural}, LipMLP~\cite{liu2022learning} and our method together with the ground truth (GT) meshes (bottom row). The middle steps usually have the highest error in quantitative evaluation. The figure shows that our method can still keep all details on the middle step meshes.\label{fig:fox-comparison}}{
    \includegraphics[width=\linewidth]{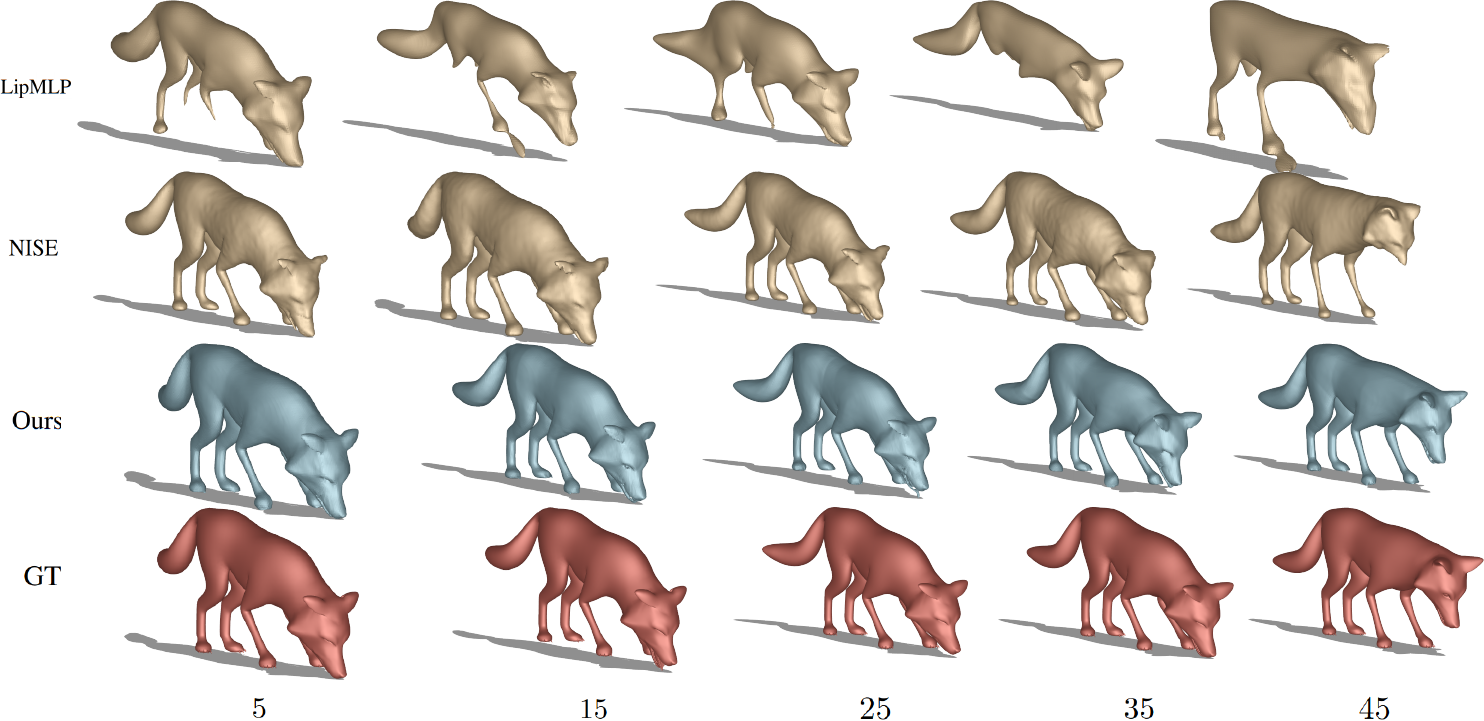}
    }
\end{minipage}%
\begin{minipage}{0.43\linewidth}
\subcaptionbox{Error plot of the fox dataset.\label{fig:fox_error}}{
    \input{figure_tex/fox_err}}
\end{minipage}
    \caption{Quantitative and qualitative evaluation of the deformed shapes. Chamfer Distance (CD) scaled by $10^3$ and Hausdorff Distance (HD) scaled by $10^2$ for the $55$ intermediate shapes.}
    % \label{fig:fox_comparison}
\end{figure}

We show the error plot for fox datasets in which the average error number is reported in~\cref{tab:comparison_error}. Even with small deformation between each key mesh, the comparison methods still report high errors on the middle step meshes. We show the ground truth meshes with the middle steps in~\cref{fig:fox-comparison} and the error plot for each mesh in~\cref{fig:fox_error}.

\subsection{Divergence-Free Constraint Ablation}
\begin{figure}[t]
    \centering
    \includegraphics[width=\linewidth]{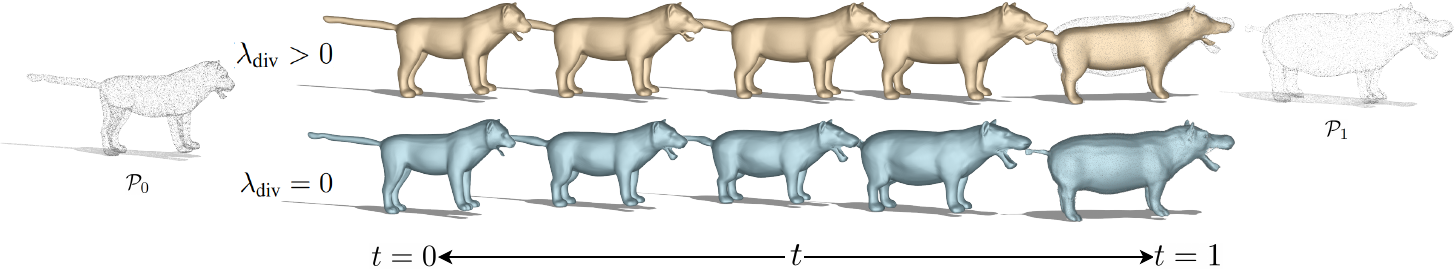}
    \caption{Full visualization of~\cref{fig:div_ablation} in main paper.}
    \label{fig:div_ablation2}
    \end{figure}

\begin{figure}[t]
    \centering
    \includegraphics[width=\linewidth]{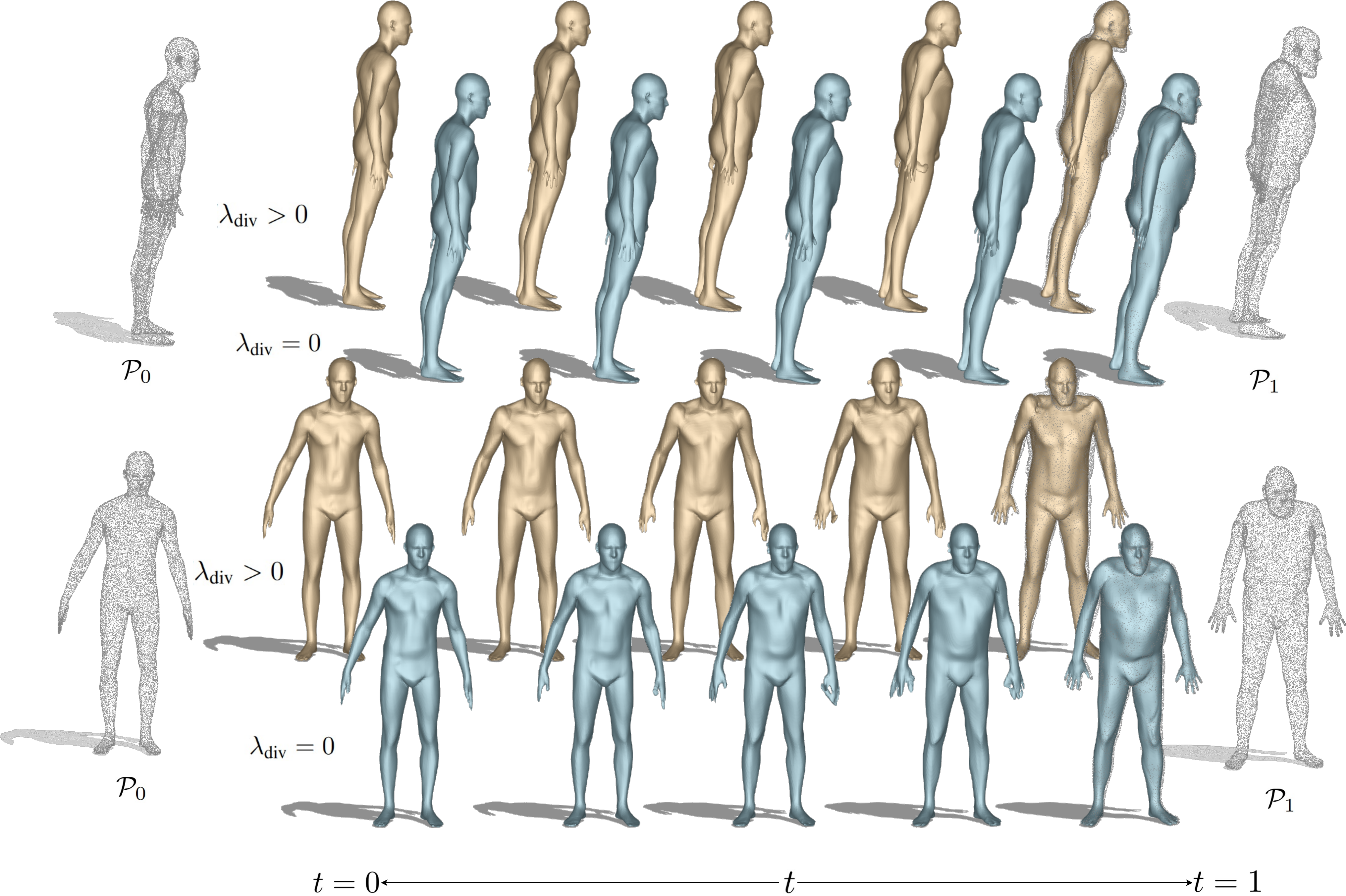}
    \caption{Visualization of divergence-free constraint on Faust dataset. With $\lambda_{\text{div}} > 0$ (yellow meshes), the deformed shapes are still slim and only adopt the movement of the target point cloud $\P_1$, while $\lambda_{\text{div}} = 0$ (blue meshes) the deformed meshes have the same gesture and body shape of the target point cloud $\P_1$}.
    \label{fig:div_ablation3}
    \end{figure}

In this section, we visualize the deformed meshes under two different settings: with divergence-free term ($\lambda_{\text{div}} > 0$) and without divergence-free term ($\lambda_{\text{div}} = 0$). In~\cref{fig:div_ablation2}, the recovered deformation meshes stay slim and thin when $\lambda_{\text{div}} > 0$ (top row) and only adopted features such as the shape of the mouth of the target point cloud. When $\lambda_{\text{div}} = 0$ (bottom row), the deformed meshes can perfectly fit the target point cloud, which means the volume expanded compared to the source point cloud.~\cref{fig:div_ablation3} shows another example of the volume persevering effect.
    
\subsection{Laplacian Constraint Ablation}
In this section, we show the visual ablation of Laplacian constraint~\eqref{eq:smoothness}. Our smoothness ablation on the velocity field ensures spatial smoothness over the integration domain, which is particularly helpful for very sparse correspondences.
\begin{figure}[h]
    \centering
    \includegraphics[width=0.9\linewidth]{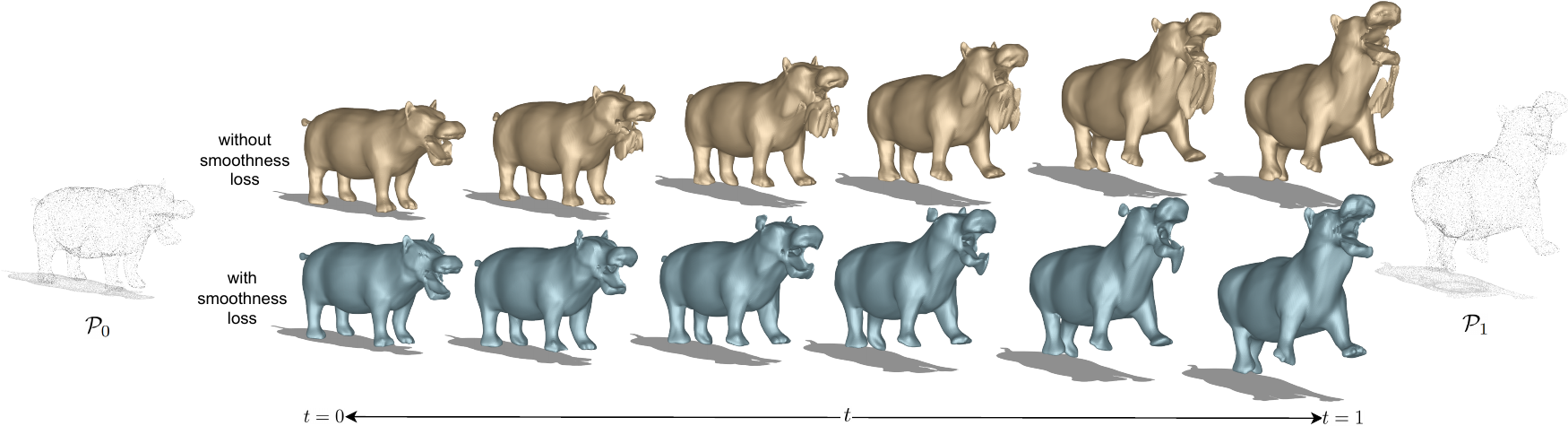}
    \caption{Our smoothness ablation on the velocity field ensures spatial smoothness over the integration domain, which is particularly helpful for very sparse correspondences.}
    \label{fig:smooth_ablation}
\end{figure}

\subsection{Modified Level Set Equation Ablation}
\begin{figure}[h]
    \centering
    \includegraphics[width=0.9\linewidth]{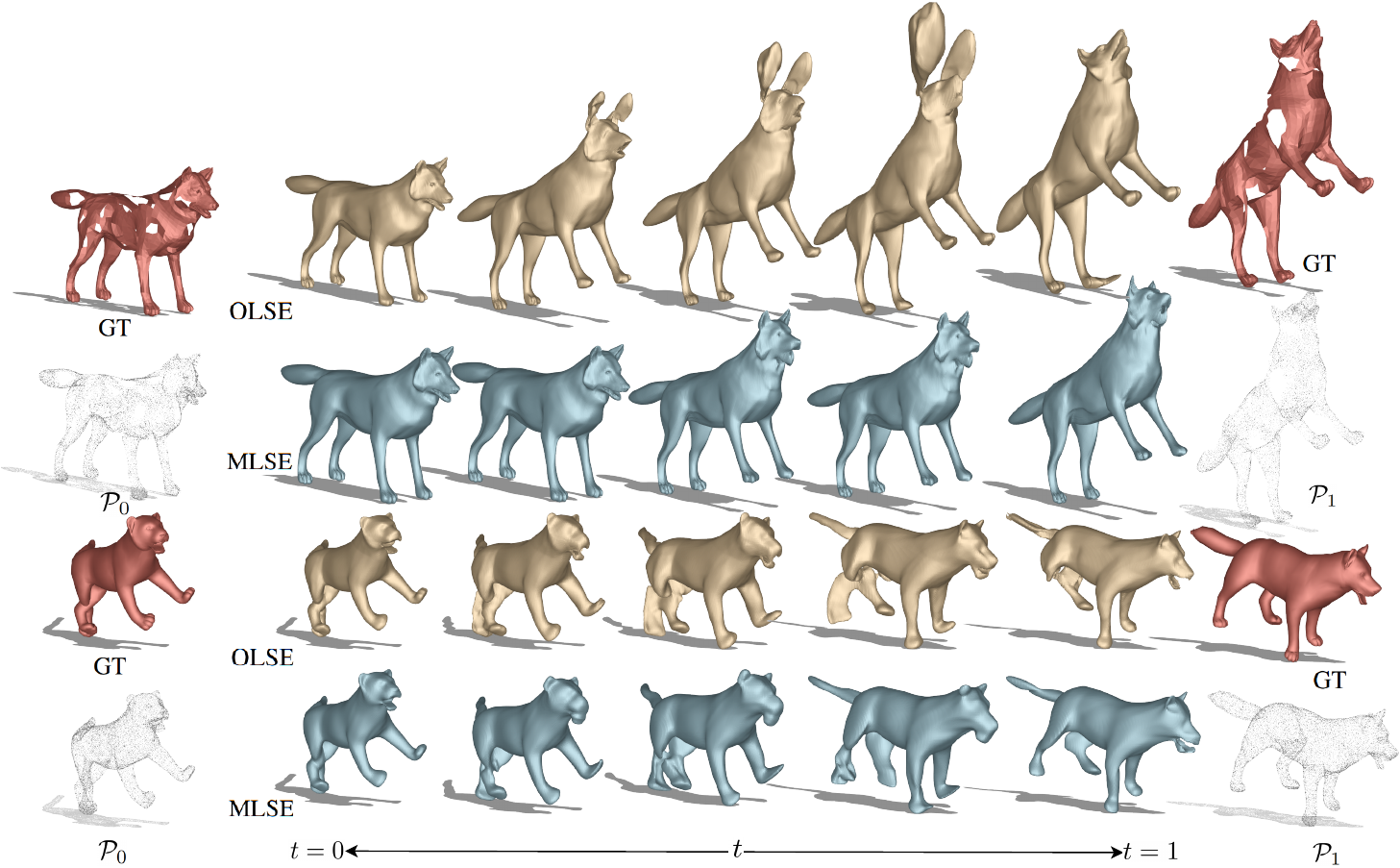}
    \caption{Additional ablation results for MLSE and OLSE.}
    \label{fig:mlse_ablation}
\end{figure}

In this section, we show additional visualization results of our proposed modified level set equation (MLSE) with original level set equation (OLSE). Comparing MLSE to OLSE, enforcing the Eikonal loss at intermediate time steps is challenging with OLSE. This process involves moving the points using velocity and then enforcing the Eikonal constraint on the moved points, which can cause a coupling effect that leads to the degeneration of the implicit field or velocity field. As shown in~\cref{fig:mlse_ablation}, the first two rows illustrate that, while the final mesh fits the target, artifacts are created in the intermediate steps. The bottom two rows demonstrate a topology change in the point cloud (e.g., the crossed legs of the cat separate later). OLSE degenerates in the middle steps, and due to the continuity of the function, it retains the degenerated legs even when fitted to the target point cloud. 

\subsection{Detail Preserving}
We show the results for meshes containing more complicated details. We deform the original Armadillo mesh using Blender to create the target shape and run our method to interpolate the intermediate shapes. As shown, our method can preserve most of the complicated geometry details.
\begin{figure}[h]
    \centering
    \includegraphics[width=\linewidth]{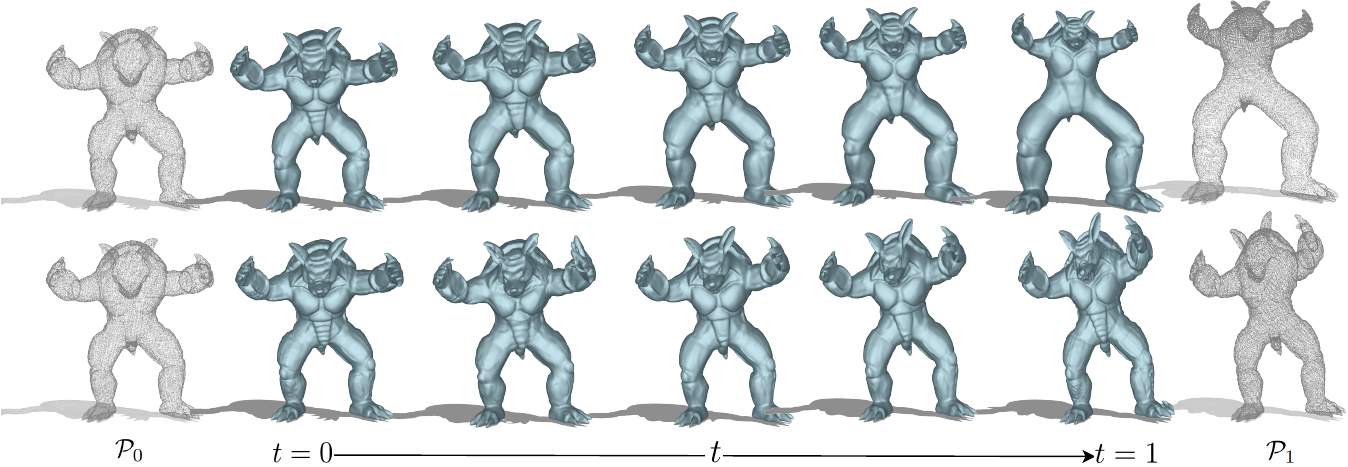}
    \caption{We show extra results on Armadillo for showing the detail-persevering of our method. Our method can preserve most of the complicated geometry details while producing physically plausible intermediate shapes.}
    \label{fig:armadillo}
\end{figure}

\subsection{Correspondence Sparsity Analysis}
\begin{figure}[t]
    \centering
    \includegraphics[width=\linewidth]{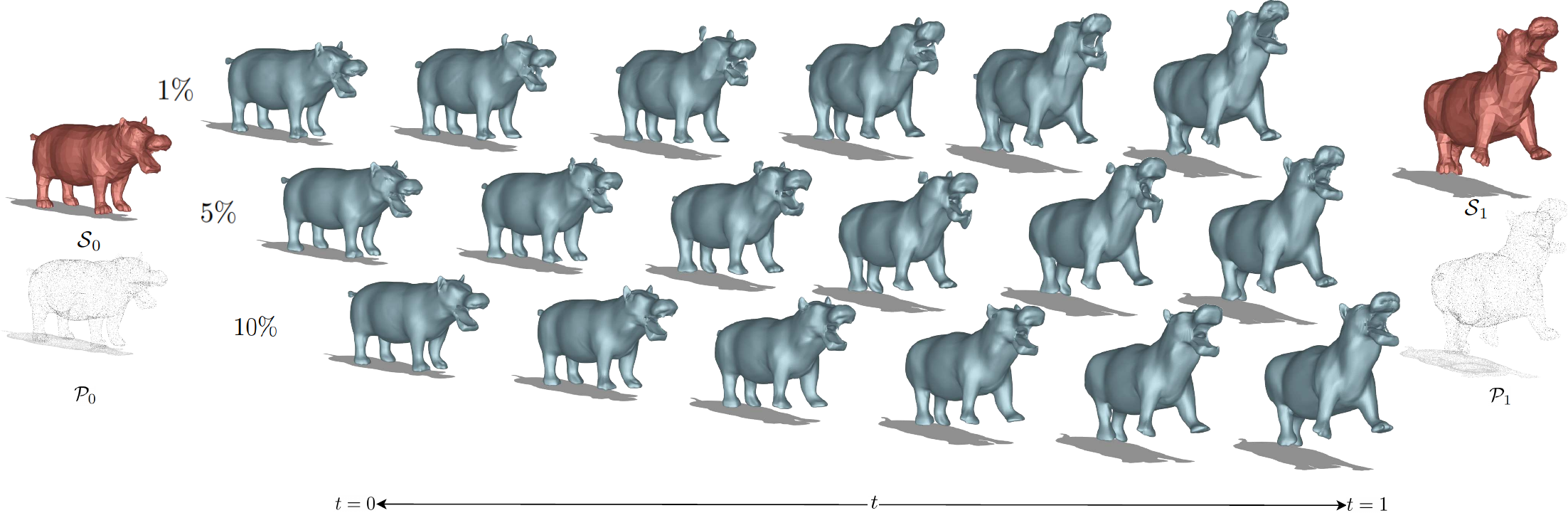}
    \caption{We explore the impact of the number of correspondences on the quality of the final deformation. When only a small percentage of correspondence is used, artifacts tend to appear in the intermediate shapes. Remarkably, our method achieves reasonable estimations with as few as approximately $1\%$ ground-truth correspondences. Furthermore, when more than approximately $10\%$ correspondences are available, our proposed method consistently delivers high-quality results.}
    \label{fig:sparsity_analysis}
\end{figure}

In this section, we present the qualitative results of our method across varying numbers of ground truth correspondences. We generated input point clouds by sampling $20,000$ points and assessed deformation quality at approximately $~1\%$, $~5\%$, and $~10\%$ correspondence levels. As illustrated in~\cref{fig:sparsity_analysis}, we display the intermediate shapes produced using different quantities of correspondences during training. Here, $\surf{0}$ and $\surf{1}$ are the ground truth meshes. $\P_0$ and $\P_1$ are the sampled point cloud inputs. Our method effectively handles different sparsity levels of correspondences and delivers high-quality results when approximately $5\%$ of the correspondences are available.

\subsection{Noisy Correspondences Analysis}
In this section, we show additional visualization of the local noise correspondence ablation together with global noisy analysis.

\begin{figure}[h]
    \centering
    \includegraphics[width=0.9\linewidth]{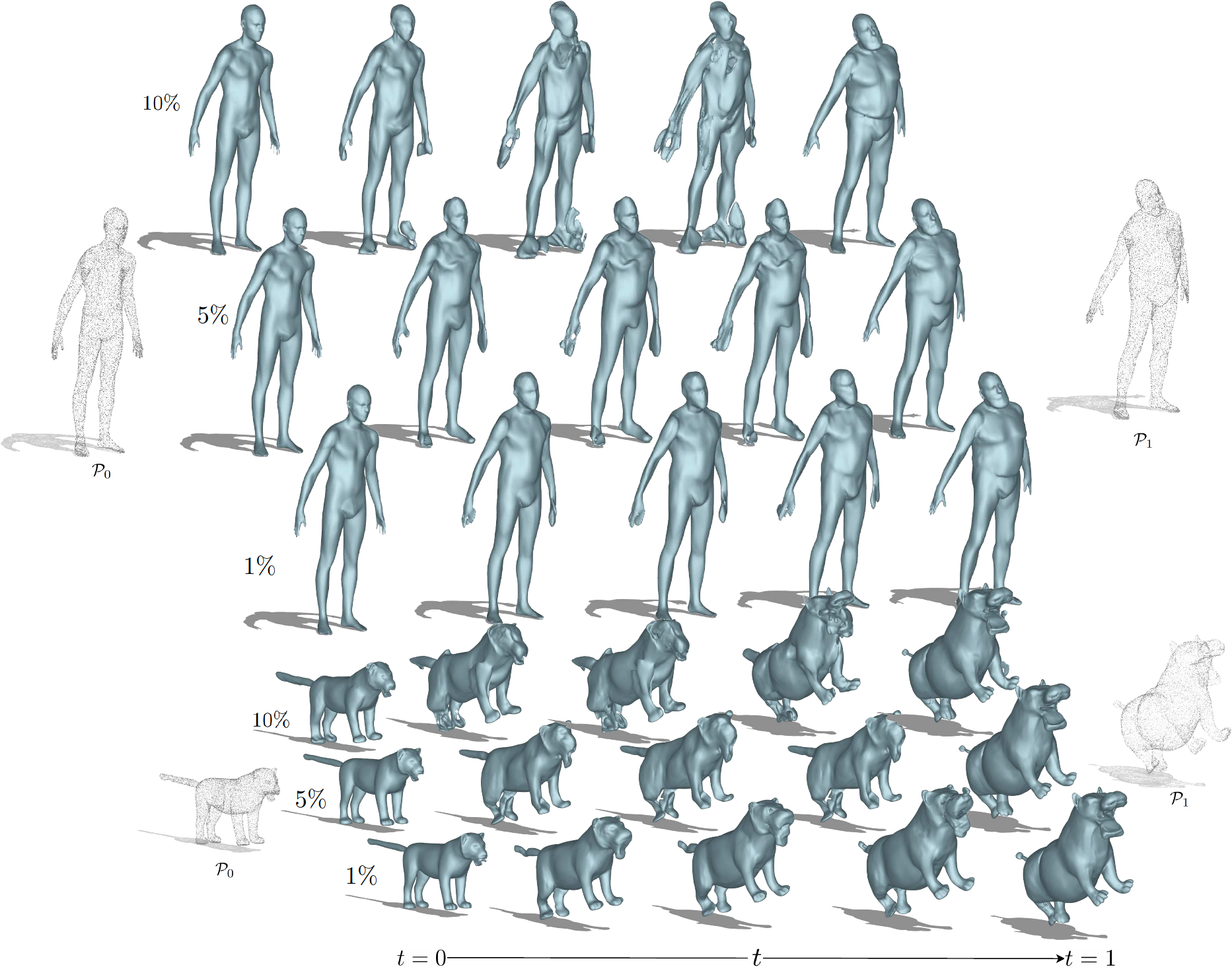}
    \caption{The input data contain around $5\%$ ground-truth correspondences. We add noise to the correspondences by randomly swapping $1\%$, $5\%$, $10\%$ of the correspondences \textbf{globally}. Qualitative results show that our method is stable up to $5\%$ error and still gives relatively reasonable results up to $10\%$ error. Note that this is an extreme situation as mismatching happens globally.}
    \label{fig:nosiy_ablation}
\end{figure}

For global noise on ground-truth correspondences. The test data contains $5\%$ correspondences relative to the total number of input points. We sample $1\%$, $5\%$, and $10\%$ of the correspondences and randomly swap them with other correspondences, regardless of whether the swapped correspondences are neighboring. This represents an extreme case for noise simulation. In this scenario, as shown in~\cref{fig:nosiy_ablation}, our method produces satisfactory results with $5\%$ wrong correspondences and still gives reasonable deformation with $10\%$ misaligned correspondences.\par
\begin{figure}[h]
    \centering
    \includegraphics[width=\linewidth]{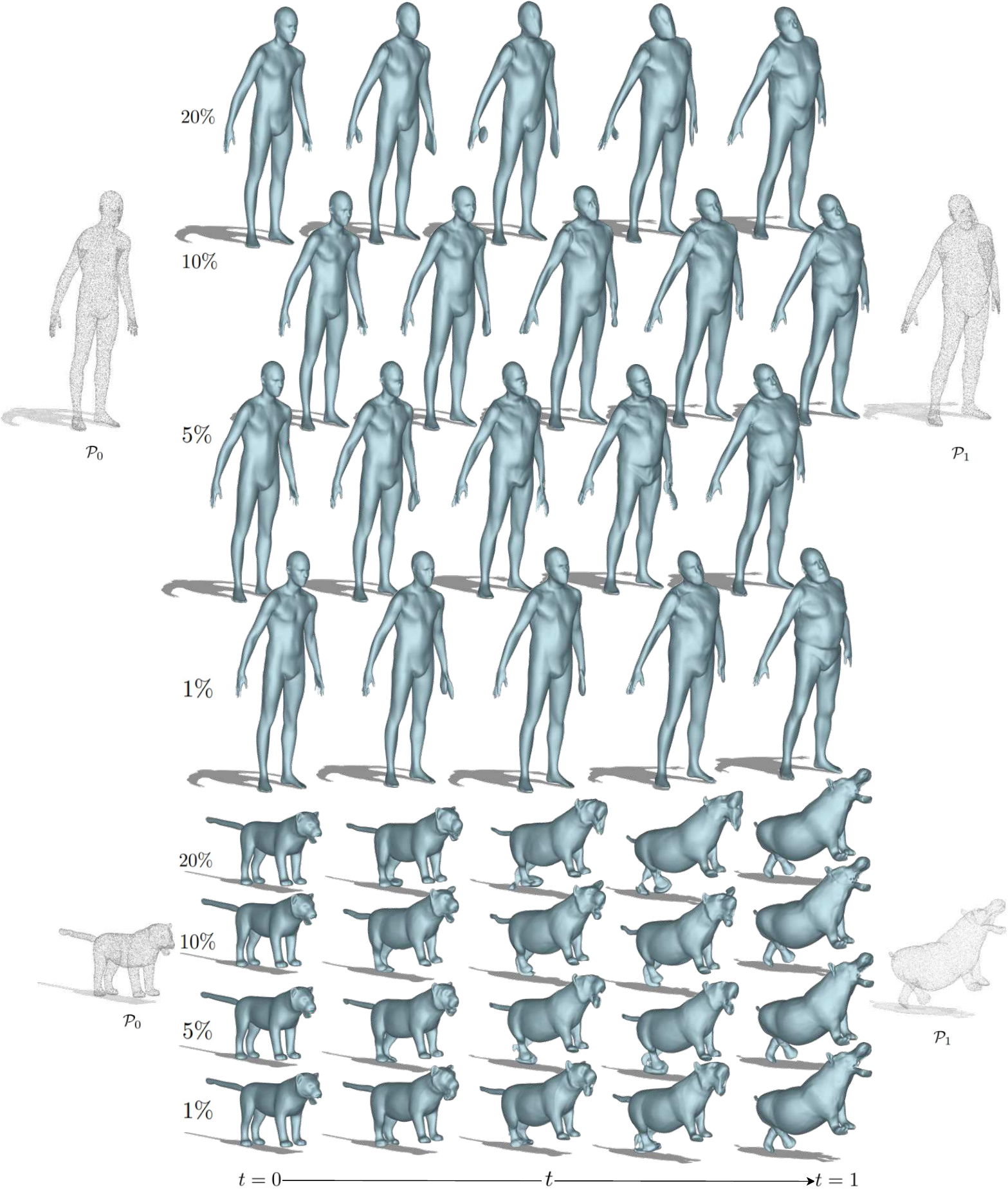}
    \caption{The input data contain around $5\%$ ground-truth correspondences. We add noise to the correspondences by randomly choosing $1\%$, $5\%$, $10\%$, $20\%$ of the GT correspondences and swap them with its $5th$ nearest neighbor correspondences. Qualitative results show that our method is stable up to $10\%$ correspondences and still gives relatively reasonable results up to $20\%$ misaligned correspondences.}
    \label{fig:nosiy_ablation22}
\end{figure}

Additionally, we show one more results on local noise ablations. The test data contains $1\%$, $5\%$ correspondences with respect to the total number of input points. We sample $5\%$, $10\%$, and $20\%$ of the correspondences and swap them with their $5$th nearest neighbor correspondences (see~\cref{fig:nosiy_ablation22}).  

\subsection{Compare With Mesh-based Methods}
\begin{figure}[t]
    \centering
    \includegraphics[width=\linewidth]{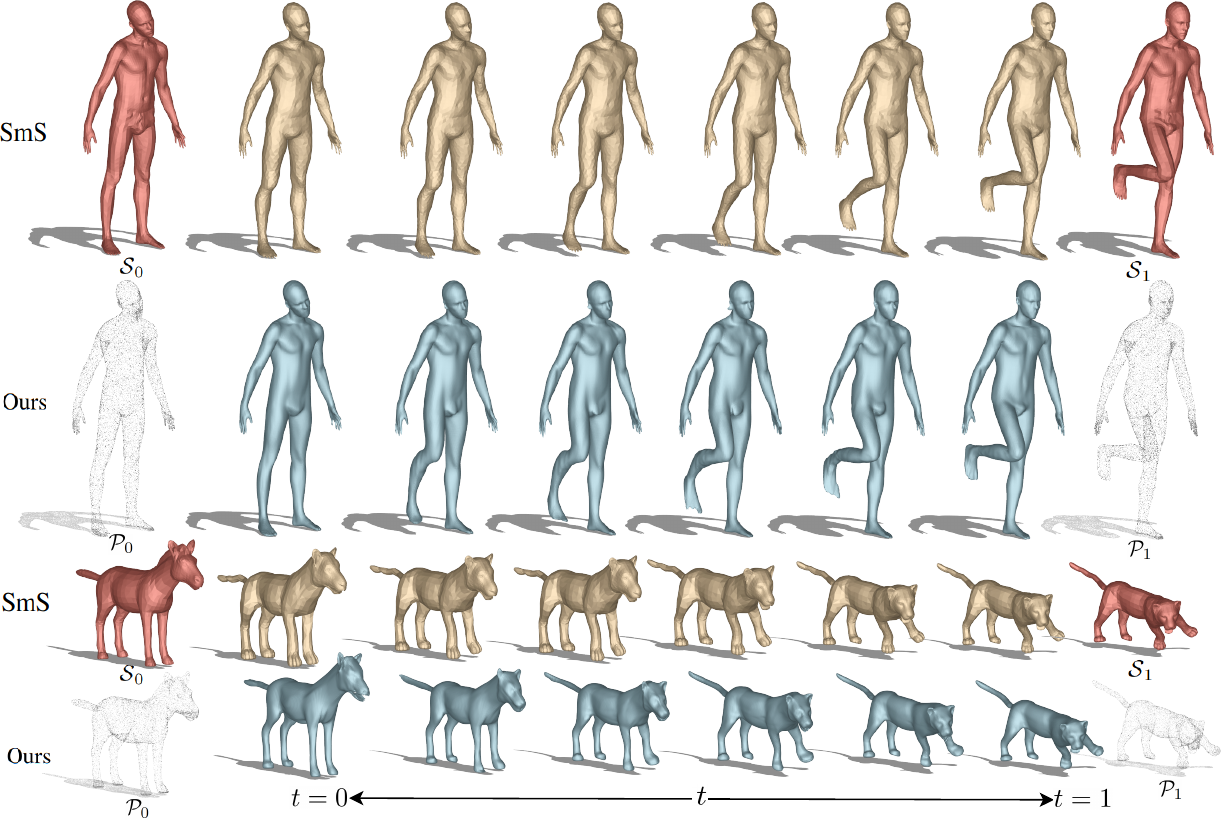}
    \caption{Comparison results against state-of-the-art mesh-based method SmS~\cite{cao2024spectral}. We produce comparable results with the meshed-based methods. However, mesh-based methods preserve more details such as human fingers. Implicit methods, on the other hand, enable rendering arbitrary resolution meshes.}
    \label{fig:mesh_based_compare1}
\end{figure}

Mesh-based surface deformation is a well-explored area, many papers have done mesh deformation with~\cite{eisenberger2020hamiltonian, alexa2023rigid, vyas2021latent} or without correspondence~\cite{eisenberger2021neuromorph, cao2024motion2vecsets, cao2024spectral}. In contrast to implicit-based methods, mesh-based methods enable more stable, artifacts-free results as no surface fitting or estimation is needed. We compare our method against the state-of-the-art mesh method, SmS \cite{cao2024spectral}. SmS does not require ground truth correspondences and is capable of producing physically plausible intermediate shapes. Our method achieves results comparable to SmS. As illustrated in~\cref{fig:mesh_based_compare1}. However, while our approach successfully preserves all the fine details, it tends to create artifacts around the surfaces and may result in less physically accurate deformations when the deformation is too large, we show some failure cases in~\cref{fig:failure_cases}. Moreover, our method reconstructs smoother meshes compared to SmS and GT meshes, because our implicit representation allows us to render higher-resolution meshes. On the contrary, mesh-based methods, like SmS keep the original resolution (same vertices, triangles, and faces) of input meshes.

\begin{figure}[t]
    \centering
    \includegraphics[width=0.9\linewidth]{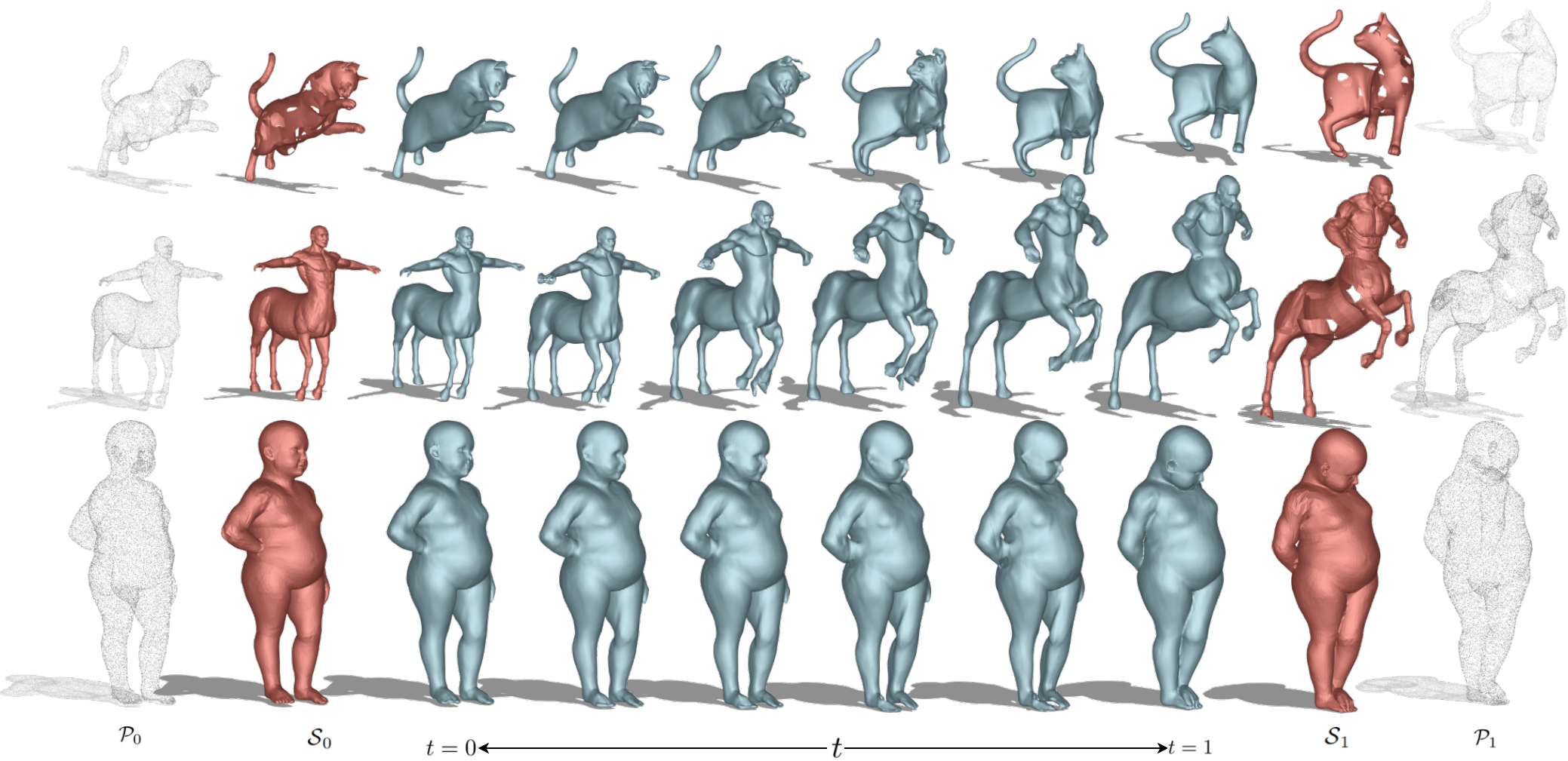}
    \caption{The proposed method can deal with inconsistent topology input, such as different incomplete shapes or self-intersect shapes.}
    \label{fig:challenge_cases}
\end{figure}

However, in some situations, such as inconsistent topology or incomplete shape without ground truth complete shape, our method can handle these challenging scenarios. Mesh-based methods struggle in these situations.~\cref{fig:challenge_cases} shows some challenging cases. In the cat (top row) example, the source and the target point cloud are sampled from meshes that have holes in the meshes. The source and target meshes are incomplete in different areas. The second example centaur shows that case of complete source mesh but incomplete target mesh. The third example even though the source and target meshes are complete, because of the overlapped feet, and overlapped arm in the target shape, the topology of the meshes is different. These three challenging examples are not feasible for the mesh-based methods. The mesh-based methods cannot handle them because the vertices and faces are not one-to-one matches anymore, even with ground truth correspondences. Since the topology of the deformed mesh is fixed, it is not trivial to deform to a shape that has a different topology. Our method, on the other hand, uses implicit representation and does not define mesh topology explicitly. Thus, our method can handle these cases.    

\subsection{Failure Cases}\label{sec:failing_cases}
\begin{figure}
    \centering
    \includegraphics[width=0.9\linewidth]{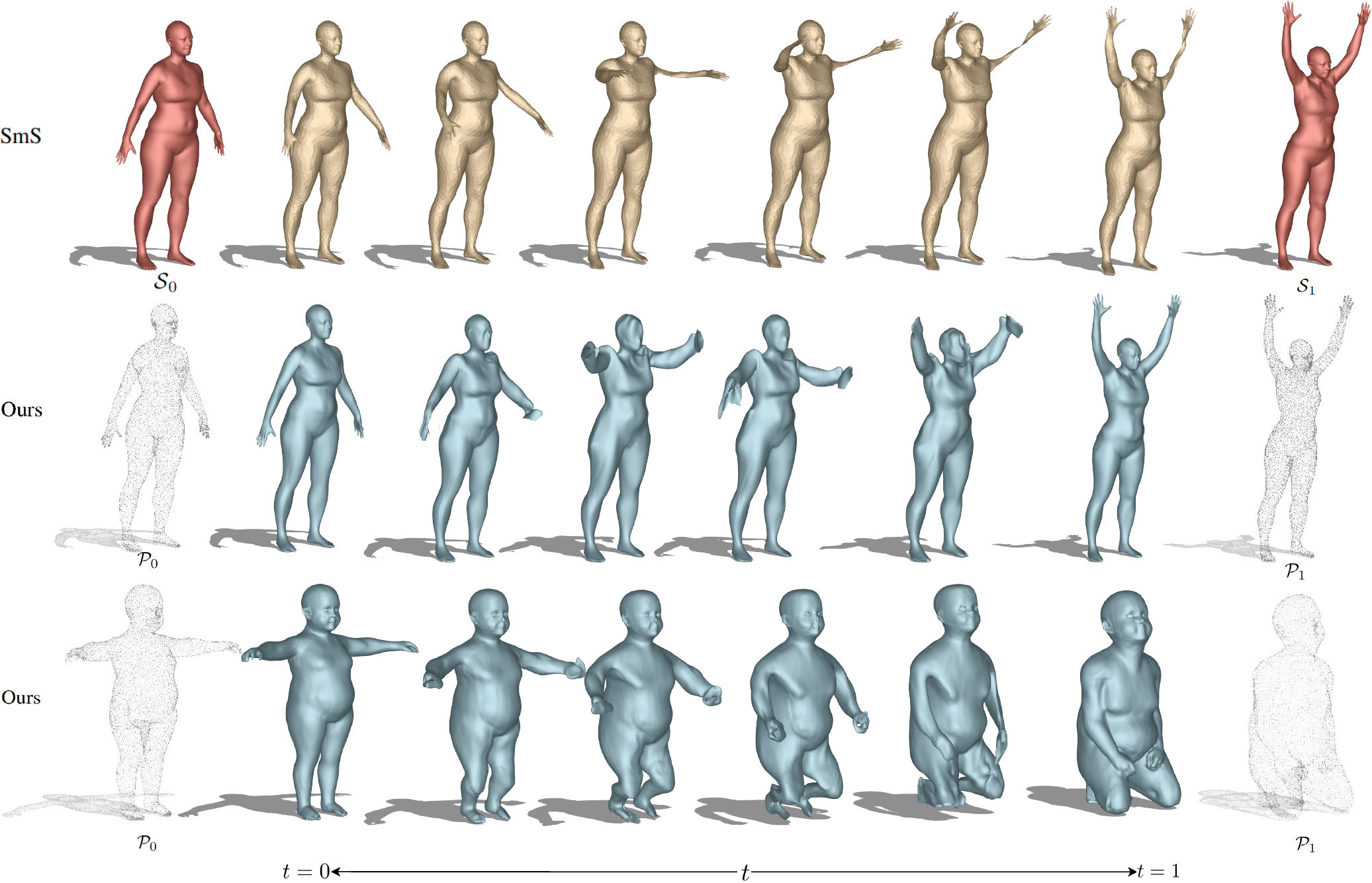}
    \caption{Failure cases of the proposed method. When the deformation is too large, our method tends to create artifacts on the surfaces.}
    \label{fig:failure_cases}
\end{figure}

As mentioned in \cref{sec:conclusion}, our work has some limitations. Here, we present some failure cases and discuss potential future improvements. Compared to mesh-based methods, our approach struggles with large deformations. \cref{fig:failure_cases} illustrates a scenario where mesh-based methods succeed, but our method produces unsatisfactory results (top two rows). In the second scenario (bottom row), although mesh-based methods fail, our method also produces artifacts around the feet due to insufficient local constraints in those areas. Another limitation occurs when there are large missing areas in the source or target point clouds. Unlike smaller holes (as shown in~\cref{fig:incomplete_shape} and~\cref{fig:challenge_cases}), substantial missing parts result in failure cases because our Velocity-Net cannot correctly move the points to the appropriate locations.

\subsection{No Correspondences}
\begin{figure} 
    \centering
    \includegraphics[width=\linewidth]{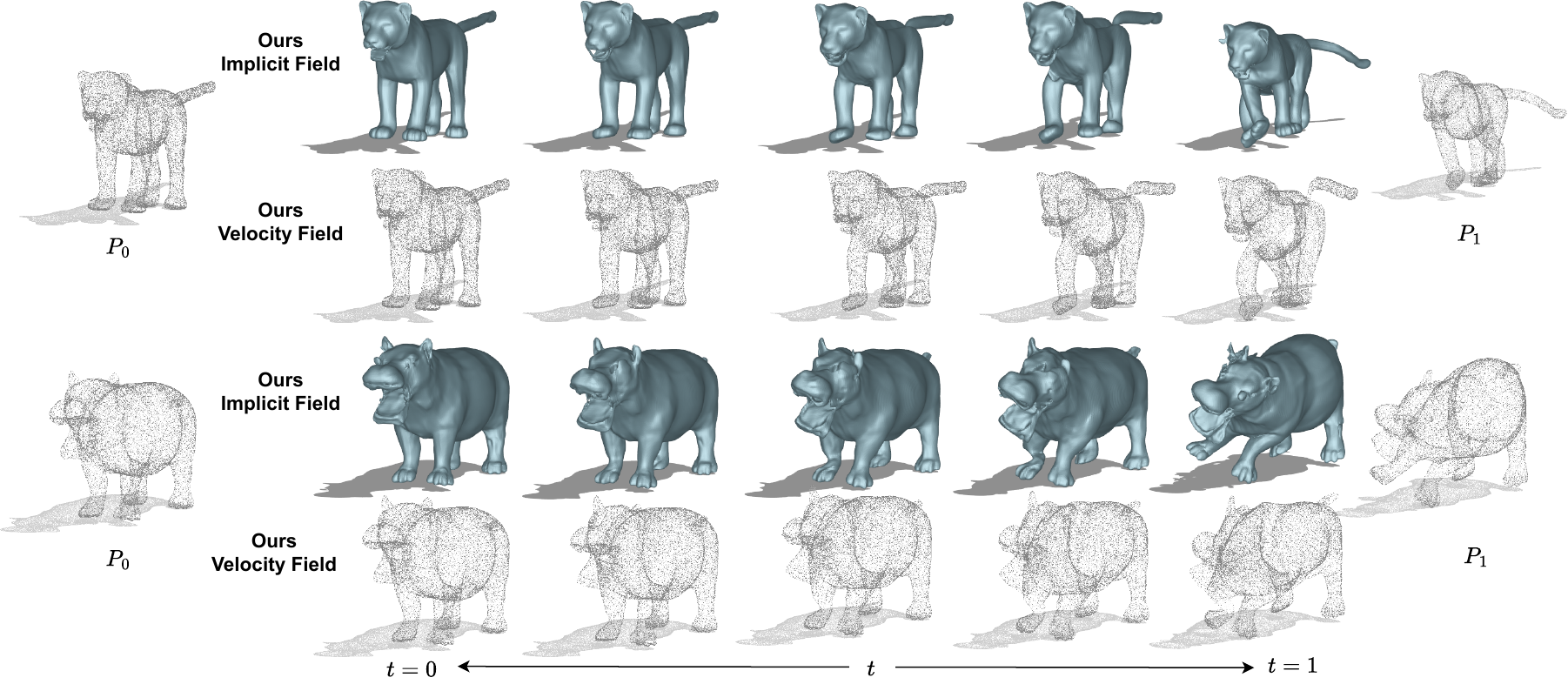}
    \caption{\textbf{No correspondences case.} In this example, we present a case with no correspondences. Remarkably, our method can still handle certain deformations, thanks to the volume-preserving constraint, smoothness constraint, and joint training with the implicit network.}
    \label{fig:smal_no_corres}
\end{figure}

In this section, we highlight scenarios where no correspondences or only partial correspondences are available. The results in~\cref{fig:smal_no_corres} demonstrate that our velocity field remains consistent, and the implicit network does not degenerate, unlike NISE~\cite{Novello2023neural} or LipMLP~\cite{liu2022learning}. Remarkably, even without correspondences, our method can still recover reasonable deformations between the two input point clouds, as shown in~\cref{fig:smal_no_corres}. We attribute this robustness to the joint training of our implicit network with strong physical constraints. However, we observe that, in the absence of correspondences, the deformations are less smooth compared to cases with correspondences, and some artifacts tend to appear around the recovered surfaces.

% \begin{figure}
%     \centering
%     \includesvg[width=\linewidth]{images/shrec16.svg}
%     \caption{\textbf{Partial shape interpolation.} We demonstrate partial correspondences using a partial target point cloud. The correspondence error is measured as the misaligned geodesic distance relative to the ground truth, with the total mesh area normalized to 1. Despite the incomplete target shape, our method successfully recovers the intermediate meshes.}
%     \label{fig:shrec16}
% \end{figure}

\end{document}